\pdfoutput=1
\documentclass[sigconf]{acmart}
\usepackage{nicefrac} 
\newcommand{\labeledNodeSet}{\mathcal{V}^{L}}
\newcommand{\labeledXSet}{\mathbf{X}^{L}}
\newcommand{\labeledYSet}{\mathbf{Y}^{L}}

\newcommand{\unLabeledNodeSet}{\mathcal{V}^{U}}
\newcommand{\unLabeledXSet}{\mathbf{X}^{U}}
\newcommand{\unLabeledYSet}{\mathbf{Y}^{U}}

\newcommand{\adagmlp}{\texttt{AdaGMLP}\@}
\usepackage{multirow}
\usepackage{float}
\usepackage{stfloats}
\usepackage{microtype}
\usepackage{graphicx}
\usepackage{subfigure}
\usepackage{booktabs} 
\usepackage{algorithmic}
\usepackage{algorithm}

\newcommand{\para}[1]{\vspace{2mm} \noindent \textbf{#1}}

\AtBeginDocument{%
  \providecommand\BibTeX{{%
    \normalfont B\kern-0.5em{\scshape i\kern-0.25em b}\kern-0.8em\TeX}}}

\setcopyright{acmcopyright}
\copyrightyear{2018}
\acmYear{2018}
\acmDOI{XXXXXXX.XXXXXXX}

\acmConference[Conference acronym 'XX]{Make sure to enter the correct
  conference title from your rights confirmation emai}{June 03--05,
  2018}{Woodstock, NY}
\acmPrice{15.00}
\acmISBN{978-1-4503-XXXX-X/18/06}

\begin{document}

\title{AdaGMLP: AdaBoosting GNN-to-MLP Knowledge Distillation}


\author{Weigang Lu}
\email{wglu@stu.xidian.edu.cn}
\orcid{https://orcid.org/0000-0003-4855-7070}
\affiliation{%
  \institution{Xidian University}
  \city{Xi'an}
  \country{China}
}

\author{Ziyu Guan}
\authornote{Corresponding Author}
\email{ziyuguan@xidian.edu.cn}
\orcid{https://orcid.org/0000-0003-2413-4698}
\affiliation{%
  \institution{Xidian University}
  \city{Xi'an}
  \country{China}
}

\author{Wei Zhao}
\email{ywzhao@mail.xidian.edu.cn}
\orcid{https://orcid.org/0000-0002-9767-1323}
\affiliation{%
  \institution{Xidian University}
  \city{Xi'an}
  \country{China}
}

\author{Yaming Yang}
\email{yym@xidian.edu.cn}
\orcid{https://orcid.org/0000-0002-8186-0648}
\affiliation{%
  \institution{Xidian University}
  \city{Xi'an}
  \country{China}
}



\begin{abstract}
Graph Neural Networks (GNNs) have revolutionized graph-based machine learning, but their heavy computational demands pose challenges for latency-sensitive edge devices in practical industrial applications. In response, a new wave of methods, collectively known as GNN-to-MLP Knowledge Distillation, has emerged. They aim to transfer GNN-learned knowledge to a more efficient MLP student, which offers faster, resource-efficient inference while maintaining competitive performance compared to GNNs. However, these methods face significant challenges in situations with insufficient training data and incomplete test data, limiting their applicability in real-world applications. To address these challenges, we propose \adagmlp, an AdaBoosting GNN-to-MLP Knowledge Distillation framework. It leverages an ensemble of diverse MLP students trained on different subsets of labeled nodes, addressing the issue of insufficient training data. Additionally, it incorporates a Node Alignment technique for robust predictions on test data with missing or incomplete features. Our experiments on seven benchmark datasets with different settings demonstrate that \adagmlp~outperforms existing G2M methods, making it suitable for a wide range of latency-sensitive real-world applications. We have submitted our code to the GitHub repository (\url{https://github.com/WeigangLu/AdaGMLP-KDD24}). 

\end{abstract}




\received{02 February 2024}
\received[revised]{17 May 2024}
\received[accepted]{17 May 2024}

\maketitle

\section{Introduction}
Graph Neural Networks (GNNs)~\cite{gcn,gat,sage,sgc,gin,appnp,skipnode,nodemixup} have revolutionized the field of graph-based machine learning, enabling state-of-the-art performance in various domains, including social networks~\cite{min2021stgsn,pcl}, recommendation systems~\cite{fan2019graph}, and bioinformatics~\cite{yang2022graph}. However, the neighbor-fetching operations in GNNs make it hard for practical industrial applications, particularly when it comes to latency constraints in numerous edge devices.

The quest for more efficient alternatives to GNNs has given rise to a new generation of methods, known as \textbf{G}raph Neural Network \textbf{to} \textbf{M}ulti-Layer Perceptrons (MLPs) \textbf{K}nowledge \textbf{D}istillation (\textbf{G2M KD}) techniques~\cite{edgefree,samlp,ffg2m,nosmog}. The primary idea is to transfer the knowledge learned by a GNN teacher into a MLP student via knowledge distillation~\cite{kd_kl}, which is graph-agnostic. G2M methods enable faster and less resource-intensive inference while maintaining competitive performance compared to GNNs.

Despite their promise, G2M KD methods face two critical challenges that restrict their real-world applicability: \textbf{insufficient training data} and \textbf{incomplete test data}. In many real-world scenarios, acquiring labeled graph data is a costly and time-consuming process and they often contain nodes with missing or incomplete features, particularly in the context of test (unseen) data. For example, in industries like finance and e-commerce, dealing with insufficient or incomplete data is a daily challenge since many customers refuse to provide (part of) their information. Ensuring the robustness of students in the presence of insufficient training data and incomplete test data is crucial for making informed decisions.

Unfortunately, the above challenges are ignored by existing G2M methods. In the insufficient training data case, traditional G2M methods employing a single MLP student can easily memorize the limited training data rather than learn general patterns from it, inducing degraded performance on test data. It is a more serious challenge on G2M than GNNs since GNNs can at least fetch neighbor information to obtain a more general picture of the graph. In the incomplete test data case, current G2M methods, which are typically designed for complete data, may struggle to make inference over the feature-missing data.

In response to these challenges, we propose \adagmlp~(AdaBoosting GNN-to-MLP Knowledge Distillation), a novel framework designed to address the limitations of existing G2M methods. It draws inspiration from ensemble learning~\cite{ensemble1,ensemble2} to leverage multiple MLP students for improved distilled knowledge via our developed AdaBoost Knowledge Distillation. Specifically, for each MLP student, we introduce a Random Classification and Node Alignment mechanism to enhance its generalization capabilities. This framework allows us to mitigate overfitting in scenarios with limited training data and ensure robust predictions on test data with missing or incomplete features. Through comprehensive experiments on seven benchmark graph datasets, we demonstrate that \adagmlp~surpasses the performance of state-of-the-art (SOTA) G2M methods across various scenarios, making it a promising solution for deploying efficient and adaptable models in real-world applications.

Our main contributions are summarized as follows:
\begin{itemize}
	\item \textbf{Tackling Real-world Challenges:} We identify two often-neglected challenges of insufficient training data and incomplete test data in current G2M KD methods and present experimental analysis in Sec.~\ref{sec:motivation}. These issues are particularly pronounced in G2M contexts, presenting a more serious challenge compared to their impact on GNNs. GNNs inherently leverage message passing to incorporate neighbor information, somewhat mitigating these issues. \adagmlp introduces innovative solutions to both issues, which are critically needed for real-world applications.
	
	\item \textbf{Novel Ensemble Architecture for G2M:} To address the above challenges, we propose \adagmlp as a novel framework consisting of Random Classification, Node Alignment, and AdaBoost Knowledge Distillation techniques. For the first time within the G2M knowledge distillation domain, our work pioneers the introduction of an ensemble architecture, making a significant departure from existing strategies focused on enhancing G2M through complex modifications or augmentations. The prior efforts, while valuable, have not ventured into establishing a generalizable G2M architecture. We have specifically tailored and extended AdaBoost for G2M, using it as a mechanism to significantly boost the generalization ability of individual MLP students.
	
	\item \textbf{Comprehensive Empirical Analysis of \adagmlp:} Extensive experiments reveal that \adagmlp~surpasses SOTA G2M methods in almost all the cases, underscoring its great effectiveness and generalization ability for practical applications. 	
\end{itemize}

\section{Related Works}
In this section, we introduce the works of transferring knowledge from a larger GNN teacher to a smaller student GNN or MLP. Specifically, we represent them as \textbf{G2G} (GNN-to-GNN) or \textbf{G2M} (GNN-to-MLP) \textbf{KD} (Knowledge Distillation), respectively.

\para{Graph-to-Graph Knowledge Distillation.} 
Prior researches~\cite{lassance2020deep,zhang2023iterative,ren2021multi,joshi2022representation,wu2022knowledge,zhang2019your,ren2021multi,gnn-sd} have primarily focused on training compact student GNNs from more expansive GNNs using KD techniques~\cite{kd_kl, kd_l2}. For example, methodologies like LSP~\cite{lsp} and TinyGNN~\cite{tinygnn} facilitate the transfer of localized structural insights from teacher GNNs to student GNNs. RDD~\cite{rdd} delves into the reliability aspects of nodes and edges to enhance the G2G KD. Although the student model used in CPF~\cite{cpf} is MLP, it additionally leverages label propagation, which still requires latency-inducing neighbor fetching. Nevertheless, these approaches still necessitate neighbor fetching, which can be impractical for applications where latency is a critical concern. 

\para{Graph-to-MLP Knowledge Distillation.} 
In response to latency concerns, recent advancements propose employing MLP students, eliminating the need for message passing during inference and showcasing competitive performance against GNN students. A pioneer work, GLNN~\cite{glnn} introduces a general G2M framework without propagations. It trains an MLP student guided by both ground-truth labels and soft labels from the GNN teacher. KRD~\cite{krd} develops a reliable sampling strategy to train MLPs with confident knowledge. Additionally, NOSMOG~\cite{nosmog} combines both structural and attribute features, which serve as inputs to MLPs, thus establishing a structure-aware MLP student. Similarly, GSDN~\cite{gsdn} introduces topological information into student training stage. Besides, FF-G2M~\cite{ffg2m} explores and provides low- and high-frequency knowledge from the graph for the student. While traditional G2M methods have made notable strides in mitigating latency concerns and enabling efficient knowledge transfer, they still exhibit certain limitations, particularly when faced with challenges related to limited training data and feature missing scenarios. We will discuss both limitations in Sec.~\ref{sec:motivation}.

\section{Preliminaries}
\para{Notions.}
We denote a graph as $\mathcal{G} = (\mathcal{V}, \mathcal{E})$, where $\mathcal{V}$ and $\mathcal{E}$ are the node set and edge set, respectively. let $N$ represent the total number of nodes. Node features are usually represented by the matrix $\mathbf{X} \in  \mathbb{R}^{N \times d}$, where each row $\mathbf{x}_{i}$ corresponds to the node $i$'s $d$-dimensional feature vector. The adjacency matrix $\mathbf{A} \in \mathbb{R}^{N \times N}$ indicates neighbor connections, where $\mathbf{A}_{ij} = 1$ if there is an edge $(i,j) \in \mathcal{E}$, and 0 otherwise. In this paper, we use capital letters to represent matrices, with corresponding lowercase letters used to denote specific rows within these matrices. For example, $\mathbf{x}_{i}$ represents the $i$-th row vector of $\mathbf{X}$.

\para{Node Classification Problem Statement.}
The label matrix is represented by $\mathbf{Y} \in \mathbb{R}^{N \times C}$ consisting of $N$ one-hot vectors, where $C$ is the number of classes. We use the superscript $L$ and $U$ to divide all the nodes into labeled ($\labeledNodeSet$, $\labeledXSet$, and $\labeledYSet$) and unlabeled parts ($\unLabeledNodeSet$, $\unLabeledXSet$, and $\unLabeledYSet$). The goal of node classification problem is to predict $\unLabeledYSet$ with $\mathbf{A}$, $\mathbf{X}$, and $\labeledYSet$ available. 

\para{Graph Neural Networks.}
Generally, most GNNs follow the message-passing scheme. That is, The representation $\mathbf{h}_{i}$ of each node $i$ undergoes iterative updates within each layer by gathering messages from its neighbors, denoted as $\mathcal{N}(i)$. In the $l$-th layer, $\mathbf{h}^{(l)}_{i}$ is computed from the representation of the previous layer through an aggregation process denoted as $\operatorname{AGGR}$, which is then followed by an $\operatorname{UPDATE}$ operation. This can be formally expressed as:
	\begin{align}
		&\mathbf{\tilde{h}}^{(l)}_{i} = \operatorname{AGGR}^{(l)}(\{\mathbf{h}^{(l-1)}_{i} : i \in \mathcal{N}(i)\}) \\
		&\mathbf{h}^{(l)}_{i} = \operatorname{UPDATE}^{(l)}(\mathbf{\tilde{h}}^{(l)}_{i}, \mathbf{h}^{(l-1)}_{i}).
	\end{align}

\para{Graph-to-MLP Knowledge Distillation.}
\cite{kd_kl} first introduces the concept of KD to enforce a simple student to mimic a more complex teacher. Notably, \cite{glnn} proposed a G2M KD framework, wherein GNNs function as teachers and MLPs serve as students. Let $\mathbf{Z}^{g} \in \mathbb{R}^{N \times C}$ and $\mathbf{Z}^{m} \in \mathbb{R}^{N \times C}$ represent the final outputs (prior to Softmax) of a GNN and an MLP, respectively. The G2M objective encompasses both the cross-entropy $\operatorname{CE}(\cdot, \cdot)$ between the predictions of the MLP and ground-truth labels:
	\begin{equation}
	\label{eq:ce}
		\mathcal{L}_{\mathrm{CE}} = \frac{1}{|\labeledNodeSet|}\sum_{i \in \labeledNodeSet} \operatorname{CE} (\sigma(\mathbf{z}^{m}_{i}), \mathbf{y}_{i}),
	\end{equation}
 as well as the KL-divergence $\mathcal{D}_{\mathrm{KL}} (\cdot, \cdot)$ calculated between the soft labels generated by the GNN and MLP:
	\begin{equation}
	\label{eq:kl}
		\mathcal{L}_{\mathrm{KL}} = \frac{1}{|\mathcal{V}|} \sum_{i 
		\in \mathcal{V}} \mathcal{D}_{\mathrm{KL}} (\sigma(\mathbf{z}^{g}_{i} / \tau), \sigma(\mathbf{z}^{m}_{i} / \tau)),
	\end{equation}
where $\sigma$ is the Softmax function and $\tau \in (0, 1]$ is the distillation temperature hyperparameter. Then, the overall objective $\mathcal{L}_{\mathrm{G2M}}$ is defined as follows:
	\begin{equation}
	\label{eq:g2m}
		\mathcal{L}_{\mathrm{G2M}} = \lambda \mathcal{L}_{\mathrm{CE}} + (1 - \lambda) \mathcal{L}_{\mathrm{KL}},
	\end{equation}
where $\lambda \in (0, 1)$ is a weighted parameter.

\section{Motivation}
\label{sec:motivation}
\subsection{Challenges in Existing G2M KD}
Recently, G2M KD methods~\cite{glnn,nosmog,krd} have demonstrated remarkable results on graph-based tasks, showcasing their superiority over traditional GNNs and G2G methods. This superiority primarily stems from their minimal inference computational overhead. However, these current G2M KD methods face significant challenges, often overlooked but highly relevant in practical scenarios:

\begin{figure}[ht]
\begin{center}
\centerline{\includegraphics[width=\columnwidth]{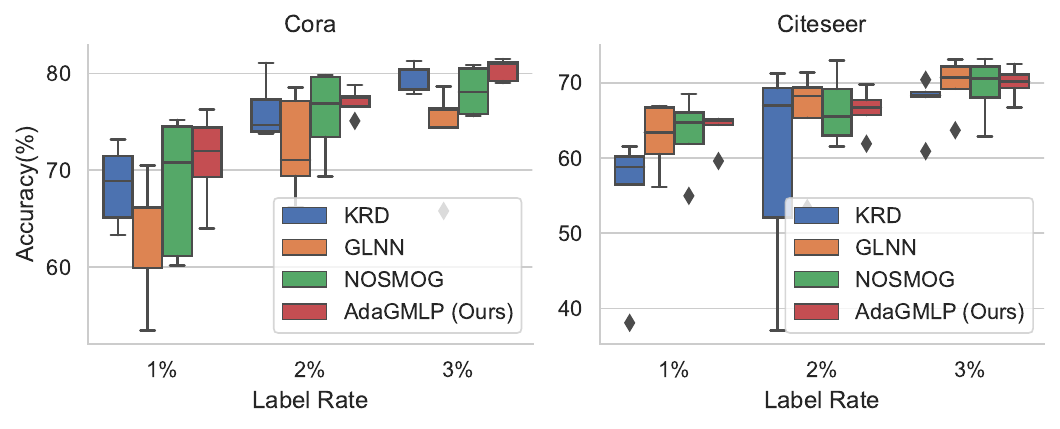}}
\caption{\textbf{[Challenge 1] Insufficient Training Data.} The single-MLP G2M methods with a single MLP student exhibit higher sensitivity to changes in label rates compared to vanilla GNNs. Notably, as the label rate decreases, there is a discernible trend of increasing box heights and the distance between outliers and box boundaries.}
\label{fig:limited_training_data}
\end{center}
\vskip -0.3in
\end{figure}

\para{[Challenge 1] Insufficient Training Data.} 
GNNs inherently possess strong generalization capabilities, benefiting from their ability to leverage unlabeled nodes via structural relationships for making predictions on unseen data.  However, transferring GNNs' knowledge into an MLP becomes problematic when training data is scarce. The first principle of G2M is \emph{``latency comes first."} Therefore, MLP sacrifices the ability of fetching neighbor information so that it can be readily applied to latency-sensitive machines. In scenarios with limited training data, relying solely on a single distilled MLP can lead to overfitting or getting stuck in local optima, resulting in inference bias. This concern motivates us to explore the generalization ability of G2M KD methods, especially in scenarios with limited data, as shown in Figure~\ref{fig:limited_training_data}. We evaluate SOTA G2M KD methods, i.e., GLNN~\cite{glnn}, KRD~\cite{krd}, and NOSMOG~\cite{nosmog} with GCN as the teacher using \texttt{Cora} and \texttt{Citeseer} datasets under varying label rates. The variability in accuracy within each method, as demonstrated by the height of the boxes and the separation between outliers and the box boundaries, reveals that the present single-MLP G2M methods exhibit higher sensitivity to changes in label rates. In contrast, our	\adagmlp~gets benefits from an AdaBoost-style ensemble and Random Classification strategy (discussed in Sec.~\ref{sec:methodlogy}) to obtain more stable performance.

\begin{figure}[ht]
\begin{center}
\centerline{\includegraphics[width=\columnwidth]{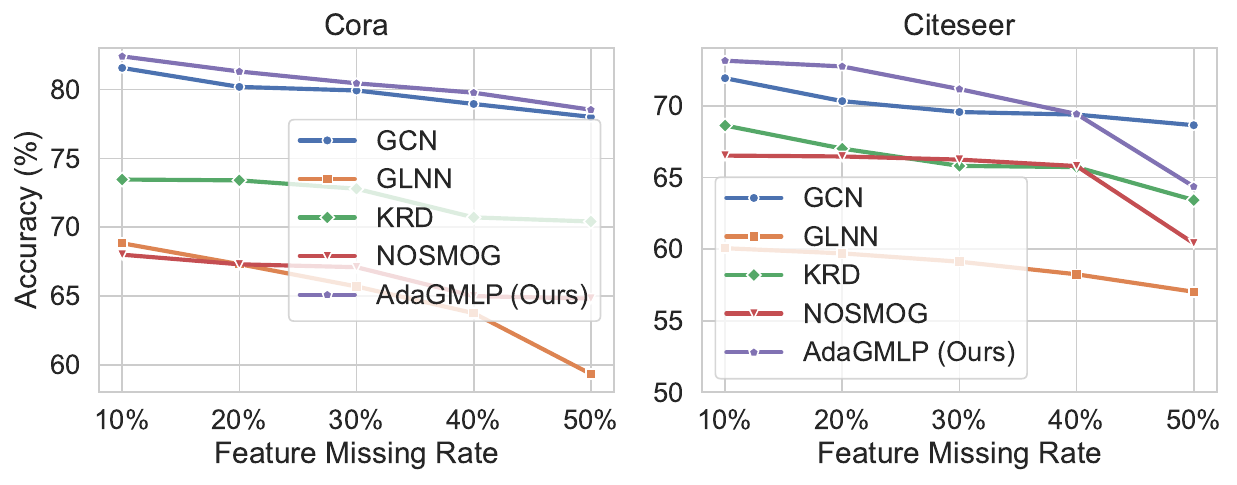}}
\caption{\textbf{[Challenge 2] Incomplete Test Data}. Traditional G2M methods suffer from performance consistent drops when more features are missing. Our \adagmlp~consistently maintains a high accuracy level, outperforming other G2M methods as the fraction of missing features increases.}
\label{fig:feat_missing}
\end{center}
\vskip -0.3in
\end{figure}

\para{[Challenge 2] Incomplete Test Data.} 
Real-world graph data is frequently incomplete, with missing features in test (new) nodes. However, traditional G2M KD methods ignore such situations and are trained under the complete datasets. When faced with feature-missing test data, they may yield suboptimal results due to the lack of mechanisms to effectively cope with this inherent incomplete features issue. This limitation becomes increasingly critical when making predictions on real-world graphs with incomplete information. In Figure~\ref{fig:feat_missing}, we visualize the performance of different G2M methods under varying levels of missing features on the \texttt{Cora} and \texttt{Citeseer} datasets. Unlike GCN that achieves a relatively stable performance, the performance of traditional G2M methods gradually decrease as more features are masked since they fail to teach the MLP student how to handle feature-missing situations. Instead, \adagmlp~tends to achieve more stable performance than counterparts due to our Node Alignment module (discussed in Sec.~\ref{sec:methodlogy}).

\subsection{Towards Addressing these Challenges}
To address these aforementioned challenges, we propose an AdaBoosting GNN-to-MLP KD (\adagmlp) framework to address these situations that impede the performance of existing G2M methods. Regarding \textbf{Challenge 1}, we tackle this by harnessing an AdaBoost-style~\cite{samme} ensemble~\cite{ensemble1, ensemble2} of multiple MLP students trained on different subsets of labeled nodes. This strategy encourages diversity in learned patterns and mitigates the risk of over-reliance on specific subsets during training. Figure~\ref{fig:limited_training_data} shows \adagmlp's great generalization ability to deal with scarce label resources. To tackle \textbf{Challenge 2}, we introduce the Node Alignment technique for each MLP student, aligning representations between labeled nodes with complete and masked features. This mechanism ensures robust predictions on test data with missing or incomplete features, thereby extending its applicability to real-world scenarios. As shown in Figure~\ref{fig:feat_missing}, \adagmlp~maintains a high and consistent accuracy level, demonstrating \adagmlp's superiority in handling feature-missing data and its potential for real-world applications.

\section{Methodology}
\label{sec:methodlogy}
\begin{figure}[!htbp]
\begin{center}
\centerline{\includegraphics[width=\columnwidth]{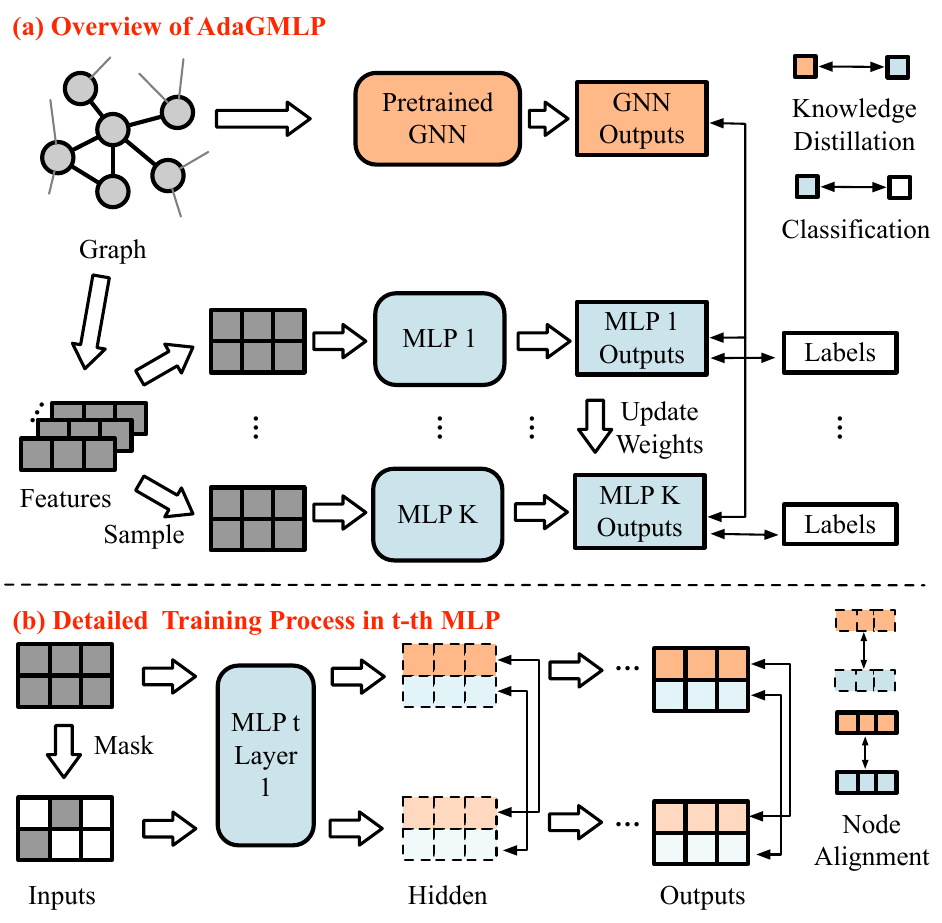}}
\caption{Illustration of \adagmlp. In (a), for each MLP, we compute the KL loss using node weights, which are determined by the difference between MLP and corresponding GNN outputs (Knowledge Distillation). Additionally, we calculate the CE loss by comparing the sampled labeled nodes with their respective ground-truth labels (Random Classification). In (b), we begin by obtain incomplete nodes with randomly masking the features of the selected nodes and inputting them into the MLP. Subsequently, we employ Mean Squared Error (MSE) loss to align their hidden representations and outputs (Node Alignment).}
\label{fig:adagmlp}
\end{center}
\vskip -0.1in
\end{figure}

In this section, we introduce \adagmlp, a methodology designed to tackle the challenges of G2M distillation while bolstering generalization and model capacity. \adagmlp~consists of a pre-trained GNN as the teacher and a compact student network with $K$ MLPs with $L$ layers. Figure~\ref{fig:adagmlp} illustrates the architecture, showcasing three fundamental components: Random Classification (RC), Node Alignment (NA), and AdaBoosting Knowledge Distillation (AdaKD).

\subsection{Random Classification}
We denote each MLP student as MS$_{1}$, MS$_{2}$, ..., MS$_{K}$. Their respective outputs are represented as $\mathbf{Z}^{m{1}}$, $\mathbf{Z}^{m_{2}}$, ..., $\mathbf{Z}^{m_{K}} \in \mathbb{R}^{N \times C}$. To enhance the student network's generalizability, we introduce randomness into the inputs for MS$_{1}$, MS$_{2}$, ..., MS$_{K-1}$ by selecting $\lfloor \nicefrac{|\labeledNodeSet|}{K} \rfloor$ nodes randomly from $\labeledNodeSet$ \emph{without replacement}. The remaining nodes are used as the input for MS$_{K}$, where $\lfloor \cdot \rfloor$ represents the floor function. Assume the labeled node subset of MS$_{k}$ is $\labeledNodeSet_{k}$, the classification objective $\mathcal{L}^{(k)}_{\mathrm{CE}}$ for MS$_{k}$ can be written as:
\begin{equation}
\label{eq:adagmlp-ce}
	\mathcal{L}^{(k)}_{\mathrm{CE}} = \frac{1}{|\labeledNodeSet_{k}|}\sum_{i \in \labeledNodeSet_{k}} \operatorname{CE} (\sigma(\mathbf{z}^{m_{k}}_{i}), \mathbf{y}_{i}).
\end{equation}
By training different MLP students on different subsets of labeled nodes, this encourages the student network to capture various patterns present in the dataset and avoids over-reliance on a specific subset of labeled nodes, leading to improved and stable performance. The Random Classification objective, $\mathcal{L}_{\mathrm{RC}}$, is presented as:
\begin{equation}
\label{eq:adagmlp-rc}
	\mathcal{L}_{\mathrm{RC}} = \frac{1}{K}\sum_{k=1}^{K} \mathcal{L}_{\mathrm{CE}}^{(k)} 
\end{equation}

\subsection{Node Alignment}
The primary idea behind Node Alignment is to align the representations of nodes with complete features (labeled nodes) and those with masked features (masked nodes) since we often encounter datasets where labeled nodes have complete feature information, while unlabeled nodes have missing features. To illustrate this, let $\mathbf{x}_{i} \in \mathbb{R}^{d}$ represent a complete node, and $\mathbf{\tilde{x}}_{i}$ signify a corrupted node with a fraction $\rho$ of its features randomly masked, where $\rho \in (0, 1)$. Consequently, we obtain outputs $\mathbf{z}^{m_{k}}_{i}$ and $\mathbf{\tilde{z}}^{m_{k}}_{i}$ as well as hidden representations $\mathbf{h}^{m_{k}, l}_{i}$ and $\mathbf{h}^{m_{k}, l}_{i}$, where $l \in \{1, 2, \cdots, L-1\}$. Now, let's delve into the two critical aspects of Node Alignment.

\para{Output Alignment (NA-O).}
In the NA-O phase, our objective is to ensure that \adagmlp's predictions on labeled nodes with completed and masked features are consistent. By minimizing the squared L2 norm between the predictions for complete and masked features, as expressed by the loss $\mathcal{L}_{\mathrm{NA\text{-}O}}$ which is expressed as:
\begin{small}
	\begin{equation}
	\label{eq:alignment-oa}
	\mathcal{L}_{\mathrm{NA\text{-}O}} = \frac{1}{K}\sum_{k=1}^{K} \mathcal{L}^{(k)}_{\mathrm{NA\text{-}O}} = \sum_{k=1}^{K} \frac{\sum_{i \in \labeledNodeSet_{k}} \lVert \mathbf{z}^{m_{k}}_{i} - \mathbf{\tilde{z}}^{m_{k}}_{i}\rVert^{2}}{K |\labeledNodeSet_{k}|}.
\end{equation}
\end{small}NA-O encourages the model to produce similar predictions for both labeled and masked nodes. This consistency contributes to stable model behavior and facilitates robust predictions.

\para{Hidden Representation Alignment (NA-H)}.
In the NA-H phase, we focus on aligning the hidden representations of nodes at different layers of the model. Similar to NA-O, we minimize the squared L2 norm between the hidden representations for complete and masked features for each layer:
\begin{small}
	\begin{equation}
	\label{eq:alignment-ha}
	\mathcal{L}_{\mathrm{NA\text{-}H}} = \frac{1}{K}\sum_{k=1}^{K} \mathcal{L}^{(k)}_{\mathrm{NA\text{-}H}} = \frac{\sum_{l=1}^{L-1} \sum_{i \in \labeledNodeSet_{k}} \lVert \mathbf{h}^{m_{k}, l}_{i} - \mathbf{\tilde{h}}^{m_{k}, l}_{i} \rVert^{2}}{K |\labeledNodeSet_{k}| (L-1)}.
	\end{equation}
\end{small}This consistency ensures that the model maintains a coherent understanding of nodes with varying feature completeness across different layers.
	
By incorporating both NA-O and NA-H with a controlling parameter $\lambda_{\mathrm{NA}} \in (0, 1)$ to optimize the following objective $\mathcal{L}_{\mathrm{NA}}$:
\begin{equation}
	\label{eq:adagmlp-na}
	\mathcal{L}_{\mathrm{NA}} = \lambda_{\mathrm{NA}} \mathcal{L}_{\mathrm{NA\text{-}O}} + (1 - \lambda_{\mathrm{NA}}) \mathcal{L}_{\mathrm{NA\text{-}H}},
\end{equation}
\adagmlp~achieves the dual goals of producing consistent predictions and maintaining coherent representations across nodes with varying feature completeness. This results in a more robust, generalizable, and stable model.

\subsection{AdaBoosting Knowledge Distillation}
We leverage AdaBoosting to obtain the collective power of multiple MLP students, further enhancing MLP students's generalization and performance. To achieve this, we adapt SAMME (Stagewise Additive Modeling using a Multi-class Exponential loss function) algorithm~\cite{samme}, which is an extension of the standard two-class AdaBoost, to propose the KD-SAMME algorithm for combining MLP students in the context of G2M. 

In KD-SAMME, we compute weighted error $e^{(k)}$, relying on KL-divergence for quantifying knowledge point (node) dissimilarity. The divergence between each node pair is denoted as $d^{(k)}_{i} = \mathcal{D}_{\mathrm{KL}} (\sigma(\mathbf{z}^{g}_{i}), \sigma(\mathbf{z}^{m_{k}}_{i}))$. The error $e^{(k)}$ is determined as:
\begin{equation}
	\label{eq:err}
	e^{(k)} =\frac{\sum_{i=1}^{N}w_{i} \left(1 - \exp(-\beta d^{(k)}_{i})\right)}{\sum_{i=1}^{N} w_{i}},
\end{equation}
where $w$ denotes the weight of $i$-th node and $\beta > 0$ controls the sensitivity to divergence between knowledge point pairs. This divergence captures the dissimilarity between individual knowledge points extracted from both the teacher and student models. 

Subsequently, we leverage this error information to compute a corresponding combining weight $\alpha^{(k)}$ for each MLP student as:
\begin{equation}
\label{eq:alpha}
	\alpha^{(k)} = \max \{\log \frac{1 - e^{(k)}}{e^{(k)}}, \epsilon\},
\end{equation}
where $\epsilon$ is an extremely small. value Further, node weights $w_{i}$ are updated by adjusting them based on $\alpha^{(k)}$ and $e^{(k)}$:

\begin{equation}
\label{eq:update-w}
w_{i} \leftarrow w_{i} \cdot \exp\left(\alpha^{(k)} \left(1 - e^{-\beta d^{(k)}_{i}}\right)\right), i=1,\cdots,N
\end{equation}

Then, node weights $w_{i}$ are normalized. The KD objective for MS$_{k}$ can be written as:
\begin{equation}
	\label{eq:adagmlp-kl}
	\mathcal{L}^{(k)}_{\mathrm{KL}} = \sum_{i 
		\in \mathcal{V}} w_{i} \mathcal{D}_{\mathrm{KL}} (\sigma(\mathbf{z}^{g}_{i} / \tau), \sigma(\mathbf{z}^{m_{k}}_{i} / \tau)).
\end{equation}

In summary, we obtain the AdaBoosting KD objective $\mathcal{L}_{\mathrm{AdaKD}}$:
\vspace{-0.5em}
\begin{equation}
	\label{eq:adagmlp-adakd}
	\mathcal{L}_{\mathrm{AdaKD}} = \frac{1}{K}\sum_{k=1}^{K}\mathcal{L}^{(k)}_{\mathrm{KL}},
\end{equation}

\begin{table*}[!htbp]
\begin{center}
\caption{Classification accuracy $\pm$ std ( \%) in the \textbf{\texttt{Transductive Setting}} and \textbf{\texttt{Inductive Setting}}.}
\label{tab:semi}
\resizebox{\textwidth}{!}{
\begin{tabular}{cc|ccccccc|c}

\toprule
\textbf{Teacher} & \textbf{Student} & \texttt{Cora} & \texttt{Citeseer} & \texttt{Pubmed} & \texttt{Photo} & \texttt{CS} & \texttt{Physics} & \texttt{ogbn-arxiv} & $\bar{\Delta}_{GLNN}$ \\ \midrule

\multicolumn{10}{c}{\textbf{\texttt{Transductive Setting}}} \\ \midrule
MLPs & - & $56.66_{\pm2.02}$ & $59.88_{\pm0.59}$ & $71.94_{\pm1.24}$ & $78.16_{\pm2.76}$ & $87.17_{\pm1.04}$ & $87.24_{\pm0.61}$ & $53.60_{\pm1.31}$ & - \\ \midrule
\multirow{5}{*}{GCN} & - & $ 82.02_{\pm0.98} $ & $71.88_{\pm0.34}$ & $77.24_{\pm0.23}$ & $90.60_{\pm2.15}$ & $89.73_{\pm0.67}$ & $92.29_{\pm0.58}$ & $ 71.22_{\pm0.18} $ & - \\
 & GLNN & $82.08_{\pm1.14} $ & $73.46_{\pm0.47}$ & $80.40_{\pm0.59}$ & $91.44_{\pm2.23}$ & $92.39_{\pm0.53}$ & $93.16_{\pm0.63}$ & $67.23_{\pm0.68}$ & - \\
 & NOSMOG & $ 82.65_{\pm1.31} $ & $73.47_{\pm1.49} $ & $ 80.95_{\pm2.21}$ & $92.39_{\pm1.95}$ & $93.71_{\pm0.63}$ & $93.49_{\pm0.42} $ & $71.07_{\pm0.24}$ & $\uparrow 1.41 \%$ \\
 & KRD  & $ 82.42_{\pm1.19} $ & $74.24_{\pm0.75}$ & $81.44_{\pm0.58}$ & $91.76_{\pm2.46}$ & $93.77_{\pm0.23}$ & $94.13_{\pm0.39}$ & $70.12_{\pm0.37}$ & $\uparrow 1.42 \%$ \\
 
 & AdaGMLP (ours)  & $ \textbf{84.26}_{\pm0.83} $ & $ \textbf{75.42}_{\pm0.39} $ & $ \textbf{81.88}_{\pm0.53} $ & $ \textbf{92.60}_{\pm0.37} $ & $ \textbf{93.79}_{\pm0.33} $ & $\textbf{94.38}_{\pm0.27}$ & $\textbf{71.45}_{\pm0.10}$ & $\uparrow \textbf{2.51} \%$ \\
 
 \midrule

\multirow{5}{*}{GraphSAGE} & - & $82.04_{\pm1.33}$ & $70.66_{\pm0.31}$ & $78.30_{\pm0.58}$ & $90.24_{\pm2.13}$ & $89.28_{\pm0.34}$ & $91.99_{\pm1.03}$ & $70.91_{\pm0.26}$ & - \\
 & GLNN & $82.24_{\pm1.11}$ & $71.90_{\pm0.76}$ & $79.78_{\pm1.46}$ & $91.44_{\pm2.23}$ & $92.86_{\pm0.28}$ & $93.28_{\pm0.94}$ & $68.63_{\pm0.12}$ & - \\
& NOSMOG & $82.74_{\pm1.53} $ & $71.95_{\pm1.39} $ & $ 80.70_{\pm1.31}$ & $92.09_{\pm1.62}$ & $93.04_{\pm0.93}$ & $93.92_{\pm1.29}$ & $70.57_{\pm0.41}$ & $\uparrow 0.89 \%$ \\
 & KRD  & $83.50_{\pm0.96}$ & $72.62_{\pm0.59}$ & $81.08_{\pm0.53}$ & $91.57_{\pm2.67}$ & $93.99_{\pm0.17} $ & $94.03_{\pm0.77}$ & $71.20_{\pm0.52}$ & $\uparrow 1.44 \%$ \\
 
 & AdaGMLP (ours)  & $ \textbf{84.10}_{\pm0.46} $ & $ \textbf{73.26}_{\pm0.29} $ & $ \textbf{81.18}_{\pm0.60} $ & $ \textbf{92.55}_{\pm2.31} $ & $ \textbf{94.06}_{\pm0.21} $ & $ \textbf{94.17}_{\pm0.57} $ & $\textbf{71.46}_{\pm0.53}$ & $\uparrow \textbf{1.93} \%$ \\

\midrule

\multirow{5}{*}{GAT} & - & $80.24_{\pm1.34} $ & $71.24_{\pm0.73}$ & $77.20_{\pm0.68}$ & $86.98_{\pm5.76}$ & $90.93_{\pm0.23}$ & $92.39_{\pm0.80}$ & $71.10_{\pm0.10}$ & - \\
 & GLNN & $81.06_{\pm1.70} $ & $69.42_{\pm3.37}$ & $80.78_{\pm0.37}$ & $86.64_{\pm9.86}$ & $93.34_{\pm0.12}$ & $93.63_{\pm0.77}$ & $68.40_{\pm0.16}$ & - \\
& NOSMOG & $81.30_{\pm1.24} $ & $70.52_{\pm1.47} $ & $80.42_{\pm2.25} $ & $92.92_{\pm1.13} $ & $94.20_{\pm0.17}$ & $93.98_{\pm0.52}$ & $71.47_{\pm0.18}$ & $\uparrow 2.07 \%$ \\
 & KRD  & $82.58_{\pm1.31}$ & $69.00_{\pm3.38}$ & $ 81.13_{\pm0.58} $ & $89.06_{\pm2.46}$ & $94.12_{\pm0.15}$ & $94.23_{\pm0.30}$ & $71.46_{\pm0.14}$ & $\uparrow 1.49 \%$ \\
 
 & AdaGMLP (ours)  & $ \textbf{83.78}_{\pm0.72} $ & $ \textbf{72.30}_{\pm1.01} $ & $ \textbf{81.68}_{\pm0.59} $ & $ \textbf{93.00}_{\pm1.62} $ & $ \textbf{94.35}_{\pm0.16} $ & $\textbf{94.33}_{\pm0.28}$ & $ \textbf{71.70}_{\pm0.33} $ & $\uparrow \textbf{3.23} \%$ \\ \midrule
 
\multicolumn{10}{c}{\textbf{\texttt{Inductive Setting}}} \\ \midrule

MLPs & - & $60.20_{\pm0.44}$ & $60.00_{\pm0.30}$ & $72.80_{\pm0.71}$ & $77.20_{
\pm3.19}$ & $ 88.97_{\pm1.12}$ & $90.16_{\pm0.42}$ & $56.39_{\pm0.56}$ & - \\ \midrule
\multirow{5}{*}{GCN} & - & $ \textbf{79.20}_{\pm0.46} $ & $\textbf{71.88}_{\pm0.36}$ & $77.36_{\pm0.71}$ & $88.67_{\pm1.22}$ & $89.55_{\pm0.48}$ & $92.47_{\pm0.45}$ & $ \textbf{70.80}_{\pm0.48}$ & - \\
 & GLNN & $72.80_{\pm0.21}$ & $70.34_{\pm0.60}$ & $78.22_{\pm0.55}$ & $88.53_{\pm2.84}$ & $91.72_{\pm0.73}$ & $93.17_{\pm0.70}$ & $61.03_{\pm0.25}$ & - \\
 & NOSMOG & $ 74.55_{\pm1.74} $ & $ 70.94_{\pm0.49} $ & $80.83_{\pm2.49} $ & $88.93_{\pm1.93}$ & $92.93_{\pm1.93}$ & $\textbf{93.97}_{\pm0.78}$ & $ 68.60_{\pm0.24}$ & $\uparrow \textbf{3.09} \%$ \\
 & KRD  & $73.52_{\pm0.21}$ & $ 70.36_{\pm0.65}$ & $80.72_{\pm1.26}$ & $ 88.16_{\pm2.02} $ & $92.09_{\pm0.61}$ & $93.79_{\pm0.48}$ & $60.41_{\pm0.26}$ & $\uparrow 0.55 \%$ \\
 
 & AdaGMLP (ours)  & $ 75.02_{\pm0.44} $ & $ 70.84_{\pm0.28} $ & $ \textbf{81.10}_{\pm0.15} $ & $ \textbf{91.15}_{\pm1.11} $ & $ \textbf{93.28}_{\pm0.28} $ & $ 93.96_{\pm0.51} $ & $64.30_{\pm0.21}$ & $\uparrow 2.62 \%$ \\
 
 \midrule

\multirow{5}{*}{GraphSAGE} & - & $ \textbf{80.32}_{\pm0.16} $ & $70.44_{\pm0.42}$ & $77.40_{\pm0.32}$ & $89.40_{\pm1.66}$ & $88.94_{\pm0.54}$ & $91.89_{\pm1.67}$ & $\textbf{70.86}_{\pm0.40}$ & - \\
 & GLNN & $70.56_{\pm1.54}$ & $70.16_{\pm1.00}$ & $79.44_{\pm1.06}$ & $88.55_{\pm2.69}$ & $91.19_{\pm0.35}$ & $92.89_{\pm1.26}$ & $61.08_{\pm0.38}$ & - \\
 & NOSMOG & $71.27_{\pm2.58} $ & $70.38_{\pm1.41}$ & $80.91_{\pm2.79}$ & $ 89.37_{\pm1.90}$ & $91.32_{\pm1.90}$ & $93.16_{\pm1.08} $ & $68.48_{\pm0.20}$ & $\uparrow 2.38 \%$ \\
 & KRD  & $70.90_{\pm1.38}$ & $70.26_{\pm0.47}$ & $80.08_{\pm0.44} $ & $89.32_{\pm1.47}$ & $92.67_{\pm0.47}$ & $93.55_{\pm1.12} $ & $61.05_{\pm0.18}$ & $\uparrow 0.65 \%$ \\
 
 & AdaGMLP (ours)  & $ 74.78_{\pm0.30} $ & $ \textbf{70.47}_{\pm0.13}$ & $ \textbf{81.34}_{\pm0.24} $ & $ \textbf{91.77}_{\pm0.43} $ & $ \textbf{93.99}_{\pm0.46} $ & $\textbf{93.98}_{\pm0.12} $ & $65.16_{\pm0.26}$ & $\uparrow \textbf{2.70} \%$ \\
 
 \midrule

\multirow{5}{*}{GAT} & - & $ \textbf{80.24}_{\pm0.90} $ & $69.72_{\pm0.61}$ & $77.00_{\pm0.68}$ & $89.97_{\pm1.86}$ & $90.22_{\pm0.94}$ & $89.95_{\pm2.29}$ & $\textbf{70.52}_{\pm0.47}$ & - \\
 & GLNN & $71.66_{\pm1.20}$ & $69.38_{\pm1.21}$ & $79.24_{\pm1.83}$ & $89.55_{\pm1.62}$ & $91.07_{\pm1.30}$ & $92.09_{\pm1.92} $ & $60.91_{\pm0.45}$ & - \\
 & NOSMOG & $72.68_{\pm2.23} $ & $70.50_{\pm2.46}$ & $81.43_{\pm3.38}$ & $89.31_{\pm1.14}$ & $91.31_{\pm1.24}$ & $93.34_{\pm1.98}$ & $68.72_{\pm0.49}$ & $\uparrow \textbf{2.85} \%$\\
 & KRD  & $71.44_{\pm1.31}$ & $69.26_{\pm1.53}$ & $ 80.52_{\pm1.36} $ & $89.49_{\pm2.85}$ & $91.68_{\pm0.36}$ & $92.83_{\pm1.38}$ & $60.95_{\pm0.60}$ & $\uparrow 0.37 \%$\\
 
 & AdaGMLP (ours)  & $ 73.92_{\pm0.68} $ & $ \textbf{71.72}_{\pm0.94} $ & $ \textbf{81.86}_{\pm0.32} $ & $ \textbf{91.44}_{\pm1.18} $ & $ \textbf{91.78}_{\pm0.75} $ & $ \textbf{93.98}_{\pm0.36} $ & $ 63.82_{\pm0.32}$ & $\uparrow 2.79 \%$ \\

\bottomrule 
\end{tabular}} 
\end{center}
\end{table*}  	

\subsection{Training and Inference}
\para{Training.}
 We define the overall \adagmlp~objective $\mathcal{L}_{\mathrm{AdaGMLP}}$ as:
\begin{equation}
	\label{eq:adagmlp}
	\mathcal{L}_{\mathrm{AdaGMLP}} = \lambda \mathcal{L}_{\mathrm{RC}} + (1 - \lambda) \mathcal{L}_{\mathrm{AdaKD}} + \mathcal{L}_{\mathrm{NA}},
\end{equation}
where $\lambda \in (0, 1)$ is a parameter to control the weight between RC and AdaKD. We also present the algorithm of \adagmlp~in Appendix~\ref{app:algorithm}.

\para{Inference.}
After the training process, we obtain a student network comprising $K$ MLPs, each associated with corresponding weights $\alpha^{(1)}, \alpha^{(2)}, \cdots, \alpha^{(K)}$. We aggregate predictions from these distinct MLPs in an AdaBoost-like manner to generate the final predicted label $\mathbf{\hat{y}}_{i}$ for the $i$-th node:
\vspace{-0.8em}
\begin{equation}
\label{eq:pred}
\mathbf{\hat{y}}_{i} = \mathop{\arg\max}\limits_{c} \sum_{k=1}^{K} \bar{\alpha}^{(k)}\sigma(\mathbf{z}^{m_{k}}_{i})
\end{equation}
Here, $\mathop{\arg\max}\limits_{c}$ denotes the selection of the class with the highest value among all classes and $\bar{\alpha}^{(k)}$ is the normalized version of $\alpha^{(k)}$. Larger $\beta$ values emphasize these under-distilled instances more, effectively making them ``stronger" in knowledge transferring.

\subsection{Complexity}
\label{sec:complexity}

\adagmlp's computational complexity primarily derives from the multiple MLPs in the ensemble and the operations involved in Node Alignment and AdaBoosting techniques. Assuming each MLP in the ensemble comprises two layers, including a transformation from $m$-dimensional input features to $d$-dimensional hidden representations and a projection from these hidden dimensions to $c$-dimensional outputs, the computational complexity for each MLP is $O(md+dc)$. For $K$ MLPs, the combined complexity for Node Alignment amounts to $O(2Kd(m+c))$. Furthermore, the AdaBoosting process, which updates weights and combines predictions across MLPs, contributes additional complexity. This aspect of the process is proportional to the number of nodes and the ensemble size, represented as $O(nK)$, where $n$ is the number of nodes. Therefore, the time complexity of our AdaGML is $O(2Kd(m + c) + nK)$ for training and $O(Kd(m + c) + nK)$ for inference.

In most cases, the hidden dimensionality $d$ often exceeds $K$, allowing \adagmlp to utilize relatively lighter MLPs with smaller $d$ while still maintaining high performance. This approach not only enhances computational efficiency but also ensures that the model remains robust and effective across various learning scenarios.

\section{Experiments}
\label{sec:exp}
In this section, we conduct a series of experiments to evaluate the performance of \adagmlp~on real-world graph datasets, addressing the following questions: 
\begin{itemize}
	\item [\textbf{Q1:}]  How does \adagmlp~perform in diverse settings (both transductive and inductive), across various real-world graphs, and with different GNN teachers (including GCN, GraphSAGE, and GAT)?
		
	\item [\textbf{Q2:}] How does \adagmlp~compare to SOTA G2M KD methods when confronted with insufficient training data?
	
	\item [\textbf{Q3:}] How effective is \adagmlp~in handling incomplete test data compared to SOTA G2M KD methods?
	
	\item [\textbf{Q4:}] Is \adagmlp~sensitive to the choice of hyper-parameters, i.e., $\lambda$, $\lambda_{\mathrm{NA}}$, $\beta$?
	
	\item [\textbf{Q5:}] How does the size of the ensemble ($K$) impact on performance?
	
	\item [\textbf{Q6:}] To what extent do the individual components of \adagmlp~contribute to its overall performance?
	
	\item [\textbf{Q7:}] Can \adagmlp~fulfill the requirements of real-world applications?
	
\end{itemize}

\subsection{Experiment Setting}
\label{sec:exp_setting}

\para{Dataset.} 
Similar to~\cite{krd}, we use six public benchmark graphs, i.e., {Cora}~\cite{cora}, \texttt{Citeseer}~\cite{citeseer}, \texttt{Pubmed}~\cite{pubmed}, \texttt{Coauthor-CS}, \texttt{Coauthor-} \texttt{Physics}, \texttt{Amazon-Photo}~\cite{coauthor}, and a large-scale graph \texttt{ogbn-arxiv}~\cite{ogb}. The statistics of datasets are provided in Appendix~\ref{app:data_stat}.

\para{Baselines.}
There are three types of baselines in this paper: \textbf{(1) GNN Teachers} including GCN~\cite{gcn}, GraphSAGE~\cite{sage}, and GAT~\cite{gat}; \textbf{(2) SOTA G2M Methods} containing GLNN~\cite{glnn}, NOSMOG~\cite{nosmog}, and KRD~\cite{krd}; \textbf{(3) SOTA G2G Methods} including CPF~\cite{cpf}, RDD~\cite{rdd}, TinyGNN~\cite{tinygnn}, GNN-SD~\cite{gnn-sd}, and LSP~\cite{lsp}. The comparison between \adagmlp~and G2G methods is described in Appendix~\ref{app:comp-g2g}.

\para{Implementation.}
The code of \adagmlp~is built on~\cite{krd} via DGL library~\cite{dgl} and we implement each MLP student with the same configuration (hidden dimensionality, number of layers) as its GNN teachers. We tune $K \in \{2, 3, 4\}$ for all the experiments except for the hyper-parameter analysis. Due to the space limitation, we present the search spaces of other hyper-parameters in the Appendix~\ref{app:imp}.


\subsection{Classification Performance Comparison (Q1)}
\label{sec:q1}
We evaluate \adagmlp~in both transductive and inductive settings, as shown in Table~\ref{tab:semi}. For all the comparing G2M methods, we evaluate the models with their released parameters. The best metrics are marked by \textbf{bold}.

In the transductive setting, \adagmlp~demonstrates superior classification accuracy compared to other G2M methods and even exceeds its GNN teachers on various datasets. In the inductive setting, \adagmlp~competes well with the SOTA methods. The average improvement over GLNN ($\bar{\Delta}_{GLNN}$), which is the representative method, varies across datasets. 

It's worth noting that \adagmlp~doesn't consistently outperform NOSMOG in some cases as it dose in the transductive setting. It is because NOSMOG benefits from access to test (unseen) node structural information, which is not typically available in real-world scenarios. Considering this, the strong performance of NOSMOG in the inductive setting should be interpreted with caution. It may not be the best choice for real-world scenarios where structural information about test nodes is unknown. In contrast, \adagmlp~performs competitively in the inductive setting without relying on any information about unseen nodes. This highlights its practical applicability and versatility, as it can handle scenarios where the test node's structure is not available, making it a more robust choice for real-world applications.

\subsection{Insufficient Training Data Case (Q2)}
\label{sec:q2}
\begin{table}[!htbp]
\begin{center}
\caption{Classification accuracy $\pm$ std (\%) in the \textbf{\texttt{Insufficient Training Data Setting}} with various label rates.}
\label{tab:insufficient}
\resizebox{1\columnwidth}{!}{
\begin{tabular}{c c c cc cc c} 
\toprule
\multirow{2}{*}{\textbf{{Dataset}}} & \textbf{Label} & \textbf{GCN} & \multicolumn{4}{c}{\textbf{Student}} \\ 
\cmidrule(lr){4-7} 
~ & \textbf{Rate} & \textbf{Teacher} & GLNN  & KRD & NOSMOG & AdaGMLP \\ \midrule

\multirow{3}{*}{\texttt{Cora}} 
 & 1\% & $67.90_{\pm4.24}$ & $63.24_{\pm5.94}$ & $68.40_{\pm4.18}$ & $68.37_{\pm7.25}$ & $\textbf{71.20}_{\pm4.22}$\\
 & 2\% & $76.81_{\pm2.15}$ & $72.48_{\pm4.68}$ & $76.18_{\pm3.08}$ & $75.84_{\pm4.45}$ & $\textbf{77.92}_{\pm1.22}$\\
 & 3\% & $79.83_{\pm1.01}$ & $74.31_{\pm4.46}$ & $79.26_{\pm1.50}$ & $78.18_{\pm2.49}$ & $\textbf{80.38}_{\pm1.05}$\\
 \midrule
 
 \multirow{3}{*}{\texttt{Citeseer}} 
& 1\% & $64.14_{\pm1.72}$ & $62.74_{\pm4.38}$ & $55.02_{\pm9.64}$ & $63.16_{\pm5.28}$ & $\textbf{63.84}_{\pm2.13}$\\
 & 2\% & $67.10_{\pm1.34}$ & $65.54_{\pm6.39}$ & $59.34_{\pm14.54}$ & $66.42_{\pm4.16}$ & $\textbf{66.46}_{\pm2.61}$\\
 & 3\% & $69.06_{\pm1.82}$ & $69.78_{\pm3.71}$ & $67.30_{\pm3.69}$ & $69.35_{\pm4.13}$ & $\textbf{69.96}_{\pm1.94}$\\

 \bottomrule 
 
\end{tabular}
}
\end{center}
\end{table}
We conducted experiments with varying label rates (1\%, 2\%, and 3\%) on the \texttt{Cora} and \texttt{Citeseer} datasets in Table~\ref{tab:insufficient}. The goal was to assess how well \adagmlp~could perform compared to GCN and other G2M methods in scenarios with limited labeled data.

Traditional G2M methods struggle to match the performance of the GCN teacher in the low-label-rate settings. This is primarily because these single-student methods might be easily over-fit to limited labeled data. As a result, they tend to show higher standard deviations compared to GCN teacher and our \adagmlp.

\adagmlp~demonstrates superior adaptability and performance in this setting. Its ability to capture and utilize information efficiently from limited labeled nodes allows it to outperform traditional G2M methods and even the GCN teacher in some cases. Additionally, \adagmlp's robustness (smaller standard deviation) across different label rates demonstrates its potential in real-world applications where obtaining a large amount of labeled data is challenging or expensive. 
 
\subsection{Incomplete Testing Data Case (Q3)}
\label{sec:q3}
In the feature-missing setting, we conducted extensive experiments on the \texttt{Cora}, \texttt{Citeseer}, and \texttt{Pubmed} datasets to evaluate the performance of \adagmlp~and compare it with GCN, and three G2M methods. We examine the impact of missing rates (10\%, 20\%, 30\%, 40\%, and 50\%) on classification accuracy. The overall results are provided in Table~\ref{tab:feat-missing}.
\begin{table}[!htbp]
\begin{center}
\caption{Classification accuracy $\pm$ std (\%) in the \textbf{\texttt{Feature-missing Setting}} with test node features randomly masked according to the missing rate.}
\label{tab:feat-missing}
\resizebox{1\columnwidth}{!}{
\begin{tabular}{c c c cc cc c}
\toprule
\multirow{2}{*}{\textbf{{Dataset}}} & \textbf{Missing} & \textbf{GCN} & \multicolumn{4}{c}{\textbf{Student}} \\ 
\cmidrule(lr){4-7} 
~ & \textbf{Rate} & \textbf{Teacher} & GLNN  & KRD & NOSMOG & AdaGMLP \\ \midrule
\multirow{5}{*}{\texttt{Cora}} 
 & 10\% & $81.58_{\pm1.42}$ & $68.84_{\pm4.71}$ & $73.47_{\pm1.43}$ & $68.01_{\pm3.97}$ & $\textbf{82.42}_{\pm0.72}$\\
 & 20\% & $81.20_{\pm1.45}$ & $67.32_{\pm2.20}$ & $73.41_{\pm1.26}$ & $67.31_{\pm4.34}$ & $\textbf{81.32}_{\pm1.49}$\\
 & 30\% & $79.94_{\pm1.99}$ & $65.70_{\pm3.26}$ & $72.80_{\pm1.93}$ & $67.10_{\pm3.66}$ & $\textbf{80.46}_{\pm1.32}$\\
 & 40\% & $78.96_{\pm2.59}$ & $63.76_{\pm3.55}$ & $70.72_{\pm1.95}$ & $65.00_{\pm7.45}$ & $\textbf{79.78}_{\pm1.73}$\\
 & 50\% & $78.02_{\pm1.73}$ & $59.34_{\pm4.07}$ & $70.42_{\pm1.83}$ & $64.82_{\pm7.14}$ & $\textbf{78.54}_{\pm1.38}$\\
 \midrule
 
 \multirow{5}{*}{\texttt{Citeseer}} 
 & 10\% & $71.92_{\pm0.84}$ & $60.06_{\pm4.04}$ & $68.62_{\pm1.42}$ & $66.52_{\pm4.30}$ & $\textbf{73.14}_{\pm0.57}$\\
 & 20\% & $70.32_{\pm0.86}$ & $59.70_{\pm3.74}$ & $67.01_{\pm1.96}$ & $66.46_{\pm4.33}$ & $\textbf{72.74}_{\pm1.25}$\\
 & 30\% & $69.56_{\pm1.91}$ & $59.12_{\pm4.13}$ & $65.80_{\pm1.83}$ & $66.24_{\pm4.73}$ & $\textbf{71.16}_{\pm1.73}$\\
 & 40\% & $69.38_{\pm1.69}$ & $58.24_{\pm3.93}$ & $65.72_{\pm0.95}$ & $65.79_{\pm4.85}$ & $\textbf{69.42}_{\pm1.24}$\\
 & 50\% & $\textbf{68.64}_{\pm1.85}$ & $57.02_{\pm4.10}$ & $63.42_{\pm1.94}$ & $60.40_{\pm3.67}$ & $64.36_{\pm1.45}$\\
 \midrule
 
 \multirow{5}{*}{\texttt{Pubmed}} 
 & 10\% & $77.22_{\pm0.52}$ & $67.64_{\pm3.60}$ & $77.94_{\pm0.73}$ & $74.19_{\pm3.41}$ & $\textbf{80.26}_{\pm0.28}$\\
 & 20\% & $76.86_{\pm0.38}$ & $67.24_{\pm3.48}$ & $77.44_{\pm0.89}$ & $73.14_{\pm3.93}$ & $\textbf{79.37}_{\pm0.43}$\\
 & 30\% & $76.34_{\pm0.35}$ & $65.34_{\pm2.57}$ & $76.60_{\pm1.02}$ & $73.09_{\pm2.23}$ & $\textbf{78.14}_{\pm1.16}$\\
 & 40\% & $76.14_{\pm0.82}$ & $64.66_{\pm4.06}$ & $75.42_{\pm1.14}$ & $72.92_{\pm3.38}$ & $\textbf{77.42}_{\pm0.56}$\\
 & 50\% & $76.05_{\pm0.98}$ & $60.01_{\pm4.55}$ & $74.31_{\pm1.47}$ & $72.43_{\pm4.29}$ & $\textbf{76.48}_{\pm1.25}$\\

\bottomrule 
 
\end{tabular} \vspace{-3em}
}
\end{center}
\end{table}
As the missing rate increases, traditional G2M methods suffer from a significant drop in accuracy. This indicates their vulnerability to missing data, limiting their practicality in real-world scenarios where data completeness cannot be guaranteed. It is mainly due to three reasons: \textbf{(1) Lack of Mechanisms for Feature-missing Data:} existing G2M methods are typically trained on complete datasets where all features are available with the lack of mechanisms to effectively cope with missing features; \textbf{(2) Limited Feature Information:} these G2M methods, relying on fixed feature vectors for prediction, cannot generalize well in the feature-missing test data.

\adagmlp~consistently outperforms other G2M methods across all missing rates and even exhibits better performance over GCN in almost all the cases. It can be attributed to the Node Alignment module that teaches each MLP student to align feature-missing nodes and complete nodes. Additionally, the AdaBoost-style ensemble approach encourages each student to collectively compensate for the missing information by aggregating diverse knowledge from different subsets, resulting in more robust predictions.This robustness demonstrates \adagmlp's ability to handle real-world scenarios with incomplete data effectively.

\subsection{Hyper-parameters Analysis (Q4)}
\label{sec:q4}
In this section, we provide comprehensive analysis on \texttt{Cora} dataset to probe into three hyper-parameters in \adagmlp, i.e., balance weight $\lambda$ of $\mathcal{L}_{\mathrm{RC}}$ and $\mathcal{L}_{\mathrm{AdaKD}}$, balance weight $\lambda_{\mathrm{NA}}$ of NA-H and NA-O, and sensitivity weight $\beta$ of knowledge point pairs divergence. To obtain more focused analysis, we remove the Node Alignment module when the interested hyper-parameter is not $\lambda_{\mathrm{NA}}$.
\begin{figure}[!htbp]
	\begin{center}
		\subfigure[$\lambda$]{\includegraphics[width=0.33\linewidth]{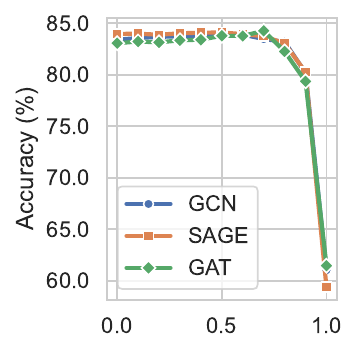} \label{fig:lamb}} \hspace{-3mm}
		\subfigure[$\lambda_{\mathrm{NA}}$]{\includegraphics[width=0.33\linewidth]{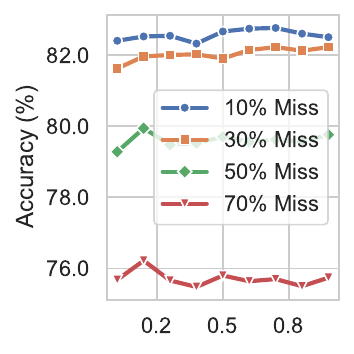}\label{fig:lamb_na}}
		\hspace{-3mm}
		\subfigure[$\beta$]{\includegraphics[width=0.33\linewidth]{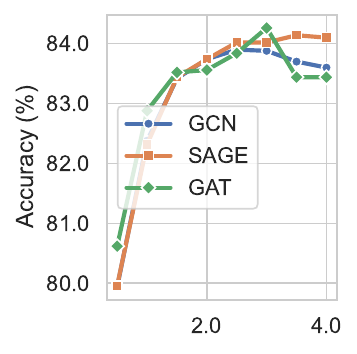} \label{fig:beta}}
	\end{center}
	\caption{Hyper-parameter Analysis on $\lambda$, $\lambda_{\mathrm{NA}}$, and $\beta$.}
	\vspace{-0.5em}
	\label{fig:hyper-parameter-study}
\end{figure}

In Figure~\ref{fig:lamb}, we tune $\lambda$ from 0 to 1 with interval of 0.1 using various GNN teachers. We can  observe a noticeable and consistent drop in performance across all teacher models when $\lambda$ exceeds a certain large threshold, e.g., 0.9.  This phenomenon can be explained by considering the role of $\lambda$ in balancing classification loss $\mathcal{L}_{\mathrm{RC}}$ and knowledge distillation loss $\mathcal{L}_{\mathrm{AdaKD}}$. When $\lambda$ is set to be very large, the model places an overwhelming emphasis on minimizing the classification error during training. This may lead to overfitting on the training data and the teacher's knowledge not being effectively distilled into the student. More interestingly, \adagmlp~maintains high performance at $\lambda = 0$ (complete knowledge distillation). It can be attributed to our AdaBoost Knowledge Distillation, which allows students to effectively transfer valuable knowledge from GNNs. 


In Figure~\ref{fig:lamb_na}, we observe a notable phenomenon:  in high feature missing settings (e.g., 70\% missing rate), smaller values of $\lambda_{\mathrm{NA}}$ lead to better results, while in low feature missing scenarios (e.g., 10\% missing rate), larger values of $\lambda_{\mathrm{NA}}$ are more effective. With a higher feature missing rate, retaining information through NA-H becomes crucial since the limited available features in test nodes can be hardly classified. Smaller $\lambda_{\mathrm{NA}}$ values emphasize NA-H and allow the model to focus more on preserving hidden representations, which are essential in recovering information from incomplete features, thereby obtaining higher performance. Conversely, with a substantial portion of features available, there is ample feature information available for most nodes. Consequently, the model can exploit this rich data to generate meaningful outputs. A larger $\lambda_{\mathrm{NA}}$ value allocates more importance to the alignment of nodes based on their output representations, which acts like consistency regularization over label information~\cite{zhu2003semi,gcn}, to obtain more robust predictions.

In Figure~\ref{fig:beta}, we explore the sensitivity of different teacher models to varying values of $\beta$ from 0.5 to 4 with interval of 0.5. The parameter $\beta$ plays a significant role in AdaBoost Knowledge Distillation, as it controls the importance of individual instances in the ensemble. Larger $\beta$ values make student more sensitive to under-distilled node pairs. The analysis suggest that we should avoid using extremely small $\beta$. 

\subsection{Ensemble Size Analysis (Q5)}
\label{sec:q5}
\begin{figure}[ht]
\begin{center}
\centerline{\includegraphics[width=\columnwidth]{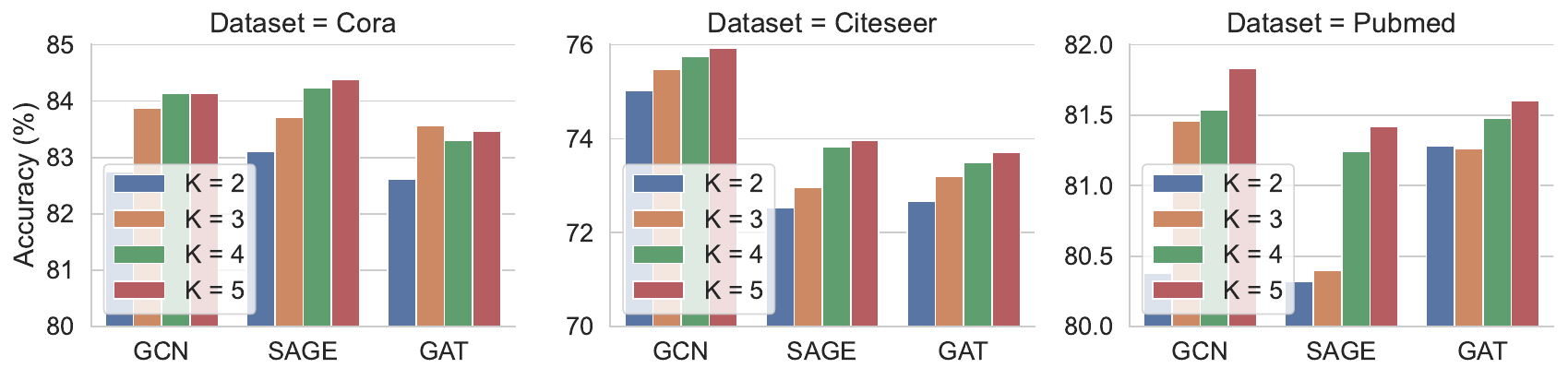}}
\caption{Ensemble Size ($K$) Analysis.}
\label{fig:K}
\end{center}
\vspace{-0.8em}
\end{figure}

In this ensemble size ($K$) ablation experiment conducted using \adagmlp~across various datasets and teacher models, we aim to explore the sensitivity of $K$ to model performance. The results reveal following noteworthy insights.

Across all datasets and teacher models, we observe that as $K$ increases, the classification accuracy generally improves. This suggests that increasing the ensemble size contributes positively to the model's performance. However, it's essential to note that the improvement tends to saturate as $K$ becomes larger. It indicates that there is an optimal point beyond which further increasing $K$ may not significantly benefit the model's performance. The sensitivity of $K$ to model performance suggests that \adagmlp~can benefit from larger $K$. Researchers can tailor the ensemble size based on their available computational resources and the dataset at hand. Smaller $K$ may suffice for some cases, while others may require larger ensembles to maximize accuracy.

%

%
%
%

\subsection{Ablation Study (Q6)}
\label{sec:q6}
\begin{figure}[!htbp]
	\begin{center}		
		\subfigure[Insufficient Training Data Setting]{\includegraphics[width=\columnwidth]{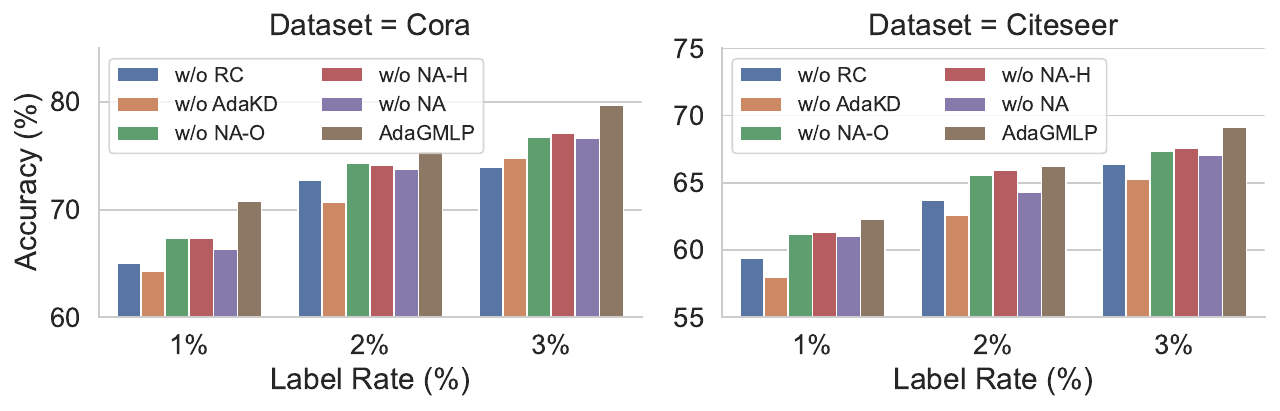} \label{fig:abl_label}}
		
		\subfigure[Incomplete Test Data Setting]{\includegraphics[width=\columnwidth]{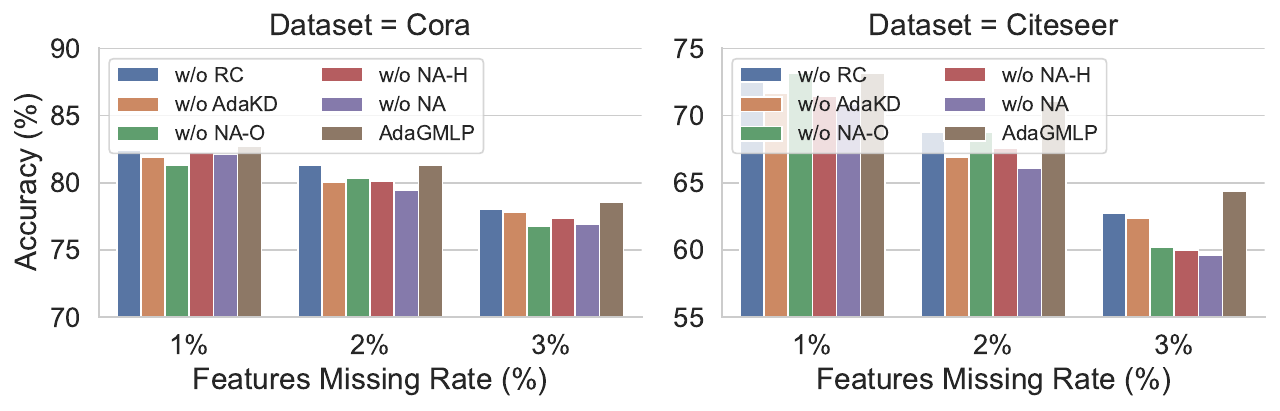}\label{fig:abl_feat}}

	\end{center}
	\caption{Ablation Study.}
	\vspace{-1.2em}
	\label{fig:abl}
\end{figure}

In this ablation experiment, we investigate the impact of different modules within \adagmlp~under two different settings, including  insufficient training data and incomplete test data. We use GCN as the teacher model and set $\lambda = 0.5, \lambda_{\mathrm{NA}}=0.5, K=2, \beta=3$ as the default setting. Different modules (RC, AdaKD, NA-O, NA-H, NA) within \adagmlp~are systematically disabled to analyze their individual contributions. The results (shown in Figure~\ref{fig:abl}) provide insights into the role of each module.

\para{Random Classification (RC).}
Removing the RC module, which involves randomly sampling training data for each student, leads to a significant drop in accuracy across all label rates and datasets. The decline is more obvious in insufficient training data setting, as shown in Figure~\ref{fig:abl_label}. This is because, in the absence of RC, students are trained on a fixed subset of data, potentially leading to overfitting. The randomness introduced by RC helps mitigate overfitting and ensures that students see diverse examples during training. It is essential for improving generalization and robustness in the presence of insufficient training data. 

\para{AdaBoosting Knowledge Distillation (AdaKD).}
Eliminating AdaKD results in a noticeable performance decrease in all the cases. AdaKD contributes to improving the student's knowledge by boosting its ability of knowledge transferring. Its role is vital for maintaining high accuracy. Moreover, it has a significant impact on performance with insufficient training data. This is because when there is limited supervision, AdaKD can help student learn from the teacher's soft labels and provide additional supervision.

\para{Node Alignment (NA).}
The NA module, formed by integrating NA-H and NA-O, is effective in maintaining model performance, especially under the incomplete test data setting, as shown in Figure~\ref{fig:abl_feat}. Removing both NA-H and NA-O leads to more pronounced performance drops, highlighting the value of their synergy within the NA module. These modules enable students to recover representations from the corrupted nodes, which is vital when dealing with incomplete test data. Without this alignment, students struggle to make predictions on unseen or partially observed nodes.

In summary, these modules serve complementary roles, and their removal impacts performance differently based on the specific challenges posed by insufficient training data or incomplete test data.

\subsection{Efficiency Analysis (Q7)}
\label{sec:q7}
\begin{figure}[!hbpt]
\vspace{-1em}
\begin{center}
\centerline{\includegraphics[width=0.9\columnwidth]{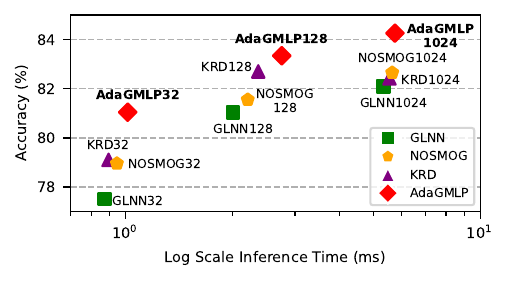}}
\caption{Accuracy \textit{vs.} Inference Time (ms).}
\label{fig:efficiency}
\end{center}
\vspace{-2em}
\end{figure}

In Figure~\ref{fig:efficiency}, we study the trade-off between performance and efficiency (inference time cost) of AdaGMLP and SOTA G2M methods on \texttt{Pubmed} dataset. For fair comparison, we use 3-layer GCN with 1024 hidden units as the teacher model and tune the hidden units in $\{32, 128, 1024\}$ for all the student model(s). We fix $K$ at 3 for \adagmlp. 

Despite a slight increase in inference time compared to other methods, \adagmlp~offers significantly better accuracy. This trade-off is often acceptable in real-world applications, where predictive performance is paramount. Besides, \adagmlp~achieves impressive accuracy (83.34$\%$) even with a relatively low hidden dimension of 128. Therefore, the increase in inference time cost due to the use of multiple MLPs is counterbalanced because of the compact student model design. The MLPs in \adagmlp are designed to be compact, with fewer parameters compared to the potentially other MLP students which demand more parameters to maintain the expressive ability. This design choice significantly reduces the computational load for each MLP. Another essential practical advantage of \adagmlp~is its inherent parallelizability. \adagmlp's architecture allows for efficient parallel computation across multiple student models. This feature can significantly reduce inference time in scenarios where parallel processing is feasible.

\section{Conclusion}
\label{sec:con}
In this work, we introduce \adagmlp, a novel ensemble framework for GNN-to-MLP Knowledge Distillation. Through an extensive series of experiments, we shed light on \adagmlp's strengths by evaluating it on various scenarios, demonstrating its great potential for real-world applications.

\begin{acks}
This work was supported in part by the National Natural Science Foundation of China under Grants 62133012, 61936006, 62073255, and 62303366, in part by the Innovation Capability Support Program of Shaanxi under Grant 2021TD-05, in part by the Fundamental Research Funds for the Central Universities under Grants QTZX23039, XJSJ23030, and in part by the Innovation Fund of Xidian University under Grant YJSJ24015.
\end{acks}

\bibliographystyle{ACM-Reference-Format}
\bibliography{refs}


\begin{thebibliography}{43}


\ifx \showCODEN    \undefined \def \showCODEN     #1{\unskip}     \fi
\ifx \showDOI      \undefined \def \showDOI       #1{#1}\fi
\ifx \showISBNx    \undefined \def \showISBNx     #1{\unskip}     \fi
\ifx \showISBNxiii \undefined \def \showISBNxiii  #1{\unskip}     \fi
\ifx \showISSN     \undefined \def \showISSN      #1{\unskip}     \fi
\ifx \showLCCN     \undefined \def \showLCCN      #1{\unskip}     \fi
\ifx \shownote     \undefined \def \shownote      #1{#1}          \fi
\ifx \showarticletitle \undefined \def \showarticletitle #1{#1}   \fi
\ifx \showURL      \undefined \def \showURL       {\relax}        \fi
\providecommand\bibfield[2]{#2}
\providecommand\bibinfo[2]{#2}
\providecommand\natexlab[1]{#1}
\providecommand\showeprint[2][]{arXiv:#2}

\bibitem[Ba and Caruana(2014)]%
        {kd_l2}
\bibfield{author}{\bibinfo{person}{Jimmy Ba} {and} \bibinfo{person}{Rich Caruana}.} \bibinfo{year}{2014}\natexlab{}.
\newblock \showarticletitle{Do deep nets really need to be deep?}
\newblock \bibinfo{journal}{\emph{Advances in neural information processing systems}}  \bibinfo{volume}{27} (\bibinfo{year}{2014}).
\newblock


\bibitem[Chen et~al\mbox{.}(2022)]%
        {samlp}
\bibfield{author}{\bibinfo{person}{Jie Chen}, \bibinfo{person}{Shouzhen Chen}, \bibinfo{person}{Mingyuan Bai}, \bibinfo{person}{Junbin Gao}, \bibinfo{person}{Junping Zhang}, {and} \bibinfo{person}{Jian Pu}.} \bibinfo{year}{2022}\natexlab{}.
\newblock \showarticletitle{SA-MLP: Distilling Graph Knowledge from GNNs into Structure-Aware MLP}.
\newblock \bibinfo{journal}{\emph{arXiv preprint arXiv:2210.09609}} (\bibinfo{year}{2022}).
\newblock


\bibitem[Chen et~al\mbox{.}(2020)]%
        {gnn-sd}
\bibfield{author}{\bibinfo{person}{Yuzhao Chen}, \bibinfo{person}{Yatao Bian}, \bibinfo{person}{Xi Xiao}, \bibinfo{person}{Yu Rong}, \bibinfo{person}{Tingyang Xu}, {and} \bibinfo{person}{Junzhou Huang}.} \bibinfo{year}{2020}\natexlab{}.
\newblock \showarticletitle{On self-distilling graph neural network}.
\newblock \bibinfo{journal}{\emph{arXiv preprint arXiv:2011.02255}} (\bibinfo{year}{2020}).
\newblock


\bibitem[Dietterich(2000a)]%
        {ensemble2}
\bibfield{author}{\bibinfo{person}{Thomas~G Dietterich}.} \bibinfo{year}{2000}\natexlab{a}.
\newblock \showarticletitle{Ensemble methods in machine learning}. In \bibinfo{booktitle}{\emph{International workshop on multiple classifier systems}}. Springer, \bibinfo{pages}{1--15}.
\newblock


\bibitem[Dietterich(2000b)]%
        {ensemble}
\bibfield{author}{\bibinfo{person}{Thomas~G Dietterich}.} \bibinfo{year}{2000}\natexlab{b}.
\newblock \showarticletitle{Ensemble methods in machine learning}. In \bibinfo{booktitle}{\emph{International workshop on multiple classifier systems}}. Springer, \bibinfo{pages}{1--15}.
\newblock


\bibitem[Fan et~al\mbox{.}(2019)]%
        {fan2019graph}
\bibfield{author}{\bibinfo{person}{Wenqi Fan}, \bibinfo{person}{Yao Ma}, \bibinfo{person}{Qing Li}, \bibinfo{person}{Yuan He}, \bibinfo{person}{Eric Zhao}, \bibinfo{person}{Jiliang Tang}, {and} \bibinfo{person}{Dawei Yin}.} \bibinfo{year}{2019}\natexlab{}.
\newblock \showarticletitle{Graph neural networks for social recommendation}. In \bibinfo{booktitle}{\emph{The world wide web conference}}. \bibinfo{pages}{417--426}.
\newblock


\bibitem[Giles et~al\mbox{.}(1998)]%
        {citeseer}
\bibfield{author}{\bibinfo{person}{C~Lee Giles}, \bibinfo{person}{Kurt~D Bollacker}, {and} \bibinfo{person}{Steve Lawrence}.} \bibinfo{year}{1998}\natexlab{}.
\newblock \showarticletitle{CiteSeer: An automatic citation indexing system}. In \bibinfo{booktitle}{\emph{Proceedings of the third ACM conference on Digital libraries}}. \bibinfo{pages}{89--98}.
\newblock


\bibitem[Hamilton et~al\mbox{.}(2017)]%
        {sage}
\bibfield{author}{\bibinfo{person}{Will Hamilton}, \bibinfo{person}{Zhitao Ying}, {and} \bibinfo{person}{Jure Leskovec}.} \bibinfo{year}{2017}\natexlab{}.
\newblock \showarticletitle{Inductive representation learning on large graphs}. In \bibinfo{booktitle}{\emph{Advances in neural information processing systems}}. \bibinfo{pages}{1024--1034}.
\newblock


\bibitem[Hastie et~al\mbox{.}(2009)]%
        {samme}
\bibfield{author}{\bibinfo{person}{Trevor Hastie}, \bibinfo{person}{Saharon Rosset}, \bibinfo{person}{Ji Zhu}, {and} \bibinfo{person}{Hui Zou}.} \bibinfo{year}{2009}\natexlab{}.
\newblock \showarticletitle{Multi-class adaboost}.
\newblock \bibinfo{journal}{\emph{Statistics and its Interface}} \bibinfo{volume}{2}, \bibinfo{number}{3} (\bibinfo{year}{2009}), \bibinfo{pages}{349--360}.
\newblock


\bibitem[Hinton et~al\mbox{.}(2015)]%
        {kd_kl}
\bibfield{author}{\bibinfo{person}{Geoffrey Hinton}, \bibinfo{person}{Oriol Vinyals}, {and} \bibinfo{person}{Jeff Dean}.} \bibinfo{year}{2015}\natexlab{}.
\newblock \showarticletitle{Distilling the knowledge in a neural network}.
\newblock \bibinfo{journal}{\emph{arXiv preprint arXiv:1503.02531}} (\bibinfo{year}{2015}).
\newblock


\bibitem[Hu et~al\mbox{.}(2020)]%
        {ogb}
\bibfield{author}{\bibinfo{person}{Weihua Hu}, \bibinfo{person}{Matthias Fey}, \bibinfo{person}{Marinka Zitnik}, \bibinfo{person}{Yuxiao Dong}, \bibinfo{person}{Hongyu Ren}, \bibinfo{person}{Bowen Liu}, \bibinfo{person}{Michele Catasta}, {and} \bibinfo{person}{Jure Leskovec}.} \bibinfo{year}{2020}\natexlab{}.
\newblock \showarticletitle{Open graph benchmark: Datasets for machine learning on graphs}.
\newblock \bibinfo{journal}{\emph{arXiv preprint arXiv:2005.00687}} (\bibinfo{year}{2020}).
\newblock


\bibitem[Joshi et~al\mbox{.}(2022)]%
        {joshi2022representation}
\bibfield{author}{\bibinfo{person}{Chaitanya~K Joshi}, \bibinfo{person}{Fayao Liu}, \bibinfo{person}{Xu Xun}, \bibinfo{person}{Jie Lin}, {and} \bibinfo{person}{Chuan~Sheng Foo}.} \bibinfo{year}{2022}\natexlab{}.
\newblock \showarticletitle{On representation knowledge distillation for graph neural networks}.
\newblock \bibinfo{journal}{\emph{IEEE Transactions on Neural Networks and Learning Systems}} (\bibinfo{year}{2022}).
\newblock


\bibitem[Kipf and Welling(2016)]%
        {gcn}
\bibfield{author}{\bibinfo{person}{Thomas~N Kipf} {and} \bibinfo{person}{Max Welling}.} \bibinfo{year}{2016}\natexlab{}.
\newblock \showarticletitle{Semi-supervised classification with graph convolutional networks}.
\newblock \bibinfo{journal}{\emph{arXiv preprint arXiv:1609.02907}} (\bibinfo{year}{2016}).
\newblock


\bibitem[Klicpera et~al\mbox{.}(2018)]%
        {appnp}
\bibfield{author}{\bibinfo{person}{Johannes Klicpera}, \bibinfo{person}{Aleksandar Bojchevski}, {and} \bibinfo{person}{Stephan G{\"u}nnemann}.} \bibinfo{year}{2018}\natexlab{}.
\newblock \showarticletitle{Predict then propagate: Graph neural networks meet personalized pagerank}.
\newblock \bibinfo{journal}{\emph{arXiv preprint arXiv:1810.05997}} (\bibinfo{year}{2018}).
\newblock


\bibitem[Lassance et~al\mbox{.}(2020)]%
        {lassance2020deep}
\bibfield{author}{\bibinfo{person}{Carlos Lassance}, \bibinfo{person}{Myriam Bontonou}, \bibinfo{person}{Ghouthi~Boukli Hacene}, \bibinfo{person}{Vincent Gripon}, \bibinfo{person}{Jian Tang}, {and} \bibinfo{person}{Antonio Ortega}.} \bibinfo{year}{2020}\natexlab{}.
\newblock \showarticletitle{Deep geometric knowledge distillation with graphs}. In \bibinfo{booktitle}{\emph{ICASSP 2020-2020 IEEE International Conference on Acoustics, Speech and Signal Processing (ICASSP)}}. IEEE, \bibinfo{pages}{8484--8488}.
\newblock


\bibitem[Lu et~al\mbox{.}(2024a)]%
        {nodemixup}
\bibfield{author}{\bibinfo{person}{Weigang Lu}, \bibinfo{person}{Ziyu Guan}, \bibinfo{person}{Wei Zhao}, \bibinfo{person}{Yaming Yang}, {and} \bibinfo{person}{Long Jin}.} \bibinfo{year}{2024}\natexlab{a}.
\newblock \showarticletitle{NodeMixup: Tackling Under-Reaching for Graph Neural Networks}.
\newblock \bibinfo{journal}{\emph{Proceedings of the AAAI Conference on Artificial Intelligence}} \bibinfo{volume}{38}, \bibinfo{number}{13} (\bibinfo{date}{Mar.} \bibinfo{year}{2024}), \bibinfo{pages}{14175--14183}.
\newblock
\urldef\tempurl%
\url{https://doi.org/10.1609/aaai.v38i13.29328}
\showDOI{\tempurl}


\bibitem[Lu et~al\mbox{.}(2023)]%
        {pcl}
\bibfield{author}{\bibinfo{person}{Weigang Lu}, \bibinfo{person}{Ziyu Guan}, \bibinfo{person}{Wei Zhao}, \bibinfo{person}{Yaming Yang}, \bibinfo{person}{Yuanhai Lv}, \bibinfo{person}{Lining Xing}, \bibinfo{person}{Baosheng Yu}, {and} \bibinfo{person}{Dacheng Tao}.} \bibinfo{year}{2023}\natexlab{}.
\newblock \showarticletitle{Pseudo contrastive learning for graph-based semi-supervised learning}.
\newblock \bibinfo{journal}{\emph{arXiv preprint arXiv:2302.09532}} (\bibinfo{year}{2023}).
\newblock


\bibitem[Lu et~al\mbox{.}(2024b)]%
        {skipnode}
\bibfield{author}{\bibinfo{person}{Weigang Lu}, \bibinfo{person}{Yibing Zhan}, \bibinfo{person}{Binbin Lin}, \bibinfo{person}{Ziyu Guan}, \bibinfo{person}{Liu Liu}, \bibinfo{person}{Baosheng Yu}, \bibinfo{person}{Wei Zhao}, \bibinfo{person}{Yaming Yang}, {and} \bibinfo{person}{Dacheng Tao}.} \bibinfo{year}{2024}\natexlab{b}.
\newblock \showarticletitle{SkipNode: On Alleviating Performance Degradation for Deep Graph Convolutional Networks}.
\newblock \bibinfo{journal}{\emph{IEEE Transactions on Knowledge and Data Engineering}} (\bibinfo{year}{2024}), \bibinfo{pages}{1--14}.
\newblock
\urldef\tempurl%
\url{https://doi.org/10.1109/TKDE.2024.3374701}
\showDOI{\tempurl}


\bibitem[McCallum et~al\mbox{.}(2000)]%
        {pubmed}
\bibfield{author}{\bibinfo{person}{Andrew~Kachites McCallum}, \bibinfo{person}{Kamal Nigam}, \bibinfo{person}{Jason Rennie}, {and} \bibinfo{person}{Kristie Seymore}.} \bibinfo{year}{2000}\natexlab{}.
\newblock \showarticletitle{Automating the construction of internet portals with machine learning}.
\newblock \bibinfo{journal}{\emph{Information Retrieval}} \bibinfo{volume}{3}, \bibinfo{number}{2} (\bibinfo{year}{2000}), \bibinfo{pages}{127--163}.
\newblock


\bibitem[Min et~al\mbox{.}(2021)]%
        {min2021stgsn}
\bibfield{author}{\bibinfo{person}{Shengjie Min}, \bibinfo{person}{Zhan Gao}, \bibinfo{person}{Jing Peng}, \bibinfo{person}{Liang Wang}, \bibinfo{person}{Ke Qin}, {and} \bibinfo{person}{Bo Fang}.} \bibinfo{year}{2021}\natexlab{}.
\newblock \showarticletitle{STGSN—A Spatial--Temporal Graph Neural Network framework for time-evolving social networks}.
\newblock \bibinfo{journal}{\emph{Knowledge-Based Systems}}  \bibinfo{volume}{214} (\bibinfo{year}{2021}), \bibinfo{pages}{106746}.
\newblock


\bibitem[Ren et~al\mbox{.}(2021)]%
        {ren2021multi}
\bibfield{author}{\bibinfo{person}{Yating Ren}, \bibinfo{person}{Junzhong Ji}, \bibinfo{person}{Lingfeng Niu}, {and} \bibinfo{person}{Minglong Lei}.} \bibinfo{year}{2021}\natexlab{}.
\newblock \showarticletitle{Multi-task Self-distillation for Graph-based Semi-Supervised Learning}.
\newblock \bibinfo{journal}{\emph{arXiv preprint arXiv:2112.01174}} (\bibinfo{year}{2021}).
\newblock


\bibitem[Sen et~al\mbox{.}(2008)]%
        {cora}
\bibfield{author}{\bibinfo{person}{Prithviraj Sen}, \bibinfo{person}{Galileo Namata}, \bibinfo{person}{Mustafa Bilgic}, \bibinfo{person}{Lise Getoor}, \bibinfo{person}{Brian Galligher}, {and} \bibinfo{person}{Tina Eliassi-Rad}.} \bibinfo{year}{2008}\natexlab{}.
\newblock \showarticletitle{Collective classification in network data}.
\newblock \bibinfo{journal}{\emph{AI magazine}} \bibinfo{volume}{29}, \bibinfo{number}{3} (\bibinfo{year}{2008}), \bibinfo{pages}{93--93}.
\newblock


\bibitem[Shchur et~al\mbox{.}(2018)]%
        {coauthor}
\bibfield{author}{\bibinfo{person}{Oleksandr Shchur}, \bibinfo{person}{Maximilian Mumme}, \bibinfo{person}{Aleksandar Bojchevski}, {and} \bibinfo{person}{Stephan G{\"u}nnemann}.} \bibinfo{year}{2018}\natexlab{}.
\newblock \showarticletitle{Pitfalls of graph neural network evaluation}.
\newblock \bibinfo{journal}{\emph{arXiv preprint arXiv:1811.05868}} (\bibinfo{year}{2018}).
\newblock


\bibitem[Tian et~al\mbox{.}(2022)]%
        {nosmog}
\bibfield{author}{\bibinfo{person}{Yijun Tian}, \bibinfo{person}{Chuxu Zhang}, \bibinfo{person}{Zhichun Guo}, \bibinfo{person}{Xiangliang Zhang}, {and} \bibinfo{person}{Nitesh Chawla}.} \bibinfo{year}{2022}\natexlab{}.
\newblock \showarticletitle{Learning mlps on graphs: A unified view of effectiveness, robustness, and efficiency}. In \bibinfo{booktitle}{\emph{The Eleventh International Conference on Learning Representations}}.
\newblock


\bibitem[Veli{\v{c}}kovi{\'c} et~al\mbox{.}(2017)]%
        {gat}
\bibfield{author}{\bibinfo{person}{Petar Veli{\v{c}}kovi{\'c}}, \bibinfo{person}{Guillem Cucurull}, \bibinfo{person}{Arantxa Casanova}, \bibinfo{person}{Adriana Romero}, \bibinfo{person}{Pietro Lio}, {and} \bibinfo{person}{Yoshua Bengio}.} \bibinfo{year}{2017}\natexlab{}.
\newblock \showarticletitle{Graph attention networks}.
\newblock \bibinfo{journal}{\emph{arXiv preprint arXiv:1710.10903}} (\bibinfo{year}{2017}).
\newblock


\bibitem[Wang et~al\mbox{.}(2019)]%
        {dgl}
\bibfield{author}{\bibinfo{person}{Minjie Wang}, \bibinfo{person}{Da Zheng}, \bibinfo{person}{Zihao Ye}, \bibinfo{person}{Quan Gan}, \bibinfo{person}{Mufei Li}, \bibinfo{person}{Xiang Song}, \bibinfo{person}{Jinjing Zhou}, \bibinfo{person}{Chao Ma}, \bibinfo{person}{Lingfan Yu}, \bibinfo{person}{Yu Gai}, \bibinfo{person}{Tianjun Xiao}, \bibinfo{person}{Tong He}, \bibinfo{person}{George Karypis}, \bibinfo{person}{Jinyang Li}, {and} \bibinfo{person}{Zheng Zhang}.} \bibinfo{year}{2019}\natexlab{}.
\newblock \showarticletitle{Deep Graph Library: A Graph-Centric, Highly-Performant Package for Graph Neural Networks}.
\newblock \bibinfo{journal}{\emph{arXiv preprint arXiv:1909.01315}} (\bibinfo{year}{2019}).
\newblock


\bibitem[Wu et~al\mbox{.}(2019)]%
        {sgc}
\bibfield{author}{\bibinfo{person}{Felix Wu}, \bibinfo{person}{Amauri Souza}, \bibinfo{person}{Tianyi Zhang}, \bibinfo{person}{Christopher Fifty}, \bibinfo{person}{Tao Yu}, {and} \bibinfo{person}{Kilian Weinberger}.} \bibinfo{year}{2019}\natexlab{}.
\newblock \showarticletitle{Simplifying graph convolutional networks}. In \bibinfo{booktitle}{\emph{International conference on machine learning}}. PMLR, \bibinfo{pages}{6861--6871}.
\newblock


\bibitem[Wu et~al\mbox{.}(2023b)]%
        {ffg2m}
\bibfield{author}{\bibinfo{person}{Lirong Wu}, \bibinfo{person}{Haitao Lin}, \bibinfo{person}{Yufei Huang}, \bibinfo{person}{Tianyu Fan}, {and} \bibinfo{person}{Stan~Z Li}.} \bibinfo{year}{2023}\natexlab{b}.
\newblock \showarticletitle{Extracting Low-/High-Frequency Knowledge from Graph Neural Networks and Injecting it into MLPs: An Effective GNN-to-MLP Distillation Framework}.
\newblock \bibinfo{journal}{\emph{arXiv preprint arXiv:2305.10758}} (\bibinfo{year}{2023}).
\newblock


\bibitem[Wu et~al\mbox{.}(2022a)]%
        {wu2022knowledge}
\bibfield{author}{\bibinfo{person}{Lirong Wu}, \bibinfo{person}{Haitao Lin}, \bibinfo{person}{Yufei Huang}, {and} \bibinfo{person}{Stan~Z Li}.} \bibinfo{year}{2022}\natexlab{a}.
\newblock \showarticletitle{Knowledge distillation improves graph structure augmentation for graph neural networks}.
\newblock \bibinfo{journal}{\emph{Advances in Neural Information Processing Systems}}  \bibinfo{volume}{35} (\bibinfo{year}{2022}), \bibinfo{pages}{11815--11827}.
\newblock


\bibitem[Wu et~al\mbox{.}(2023a)]%
        {krd}
\bibfield{author}{\bibinfo{person}{Lirong Wu}, \bibinfo{person}{Haitao Lin}, \bibinfo{person}{Yufei Huang}, {and} \bibinfo{person}{Stan~Z Li}.} \bibinfo{year}{2023}\natexlab{a}.
\newblock \showarticletitle{Quantifying the Knowledge in GNNs for Reliable Distillation into MLPs}.
\newblock \bibinfo{journal}{\emph{arXiv preprint arXiv:2306.05628}} (\bibinfo{year}{2023}).
\newblock


\bibitem[Wu et~al\mbox{.}(2022b)]%
        {gsdn}
\bibfield{author}{\bibinfo{person}{Lirong Wu}, \bibinfo{person}{Jun Xia}, \bibinfo{person}{Haitao Lin}, \bibinfo{person}{Zhangyang Gao}, \bibinfo{person}{Zicheng Liu}, \bibinfo{person}{Guojiang Zhao}, {and} \bibinfo{person}{Stan~Z Li}.} \bibinfo{year}{2022}\natexlab{b}.
\newblock \showarticletitle{Teaching Yourself: Graph Self-Distillation on Neighborhood for Node Classification}.
\newblock \bibinfo{journal}{\emph{arXiv preprint arXiv:2210.02097}} (\bibinfo{year}{2022}).
\newblock


\bibitem[Wu et~al\mbox{.}(2023c)]%
        {edgefree}
\bibfield{author}{\bibinfo{person}{Taiqiang Wu}, \bibinfo{person}{Zhe Zhao}, \bibinfo{person}{Jiahao Wang}, \bibinfo{person}{Xingyu Bai}, \bibinfo{person}{Lei Wang}, \bibinfo{person}{Ngai Wong}, {and} \bibinfo{person}{Yujiu Yang}.} \bibinfo{year}{2023}\natexlab{c}.
\newblock \showarticletitle{Edge-free but Structure-aware: Prototype-Guided Knowledge Distillation from GNNs to MLPs}.
\newblock \bibinfo{journal}{\emph{arXiv preprint arXiv:2303.13763}} (\bibinfo{year}{2023}).
\newblock


\bibitem[Xu et~al\mbox{.}(2018)]%
        {gin}
\bibfield{author}{\bibinfo{person}{Keyulu Xu}, \bibinfo{person}{Weihua Hu}, \bibinfo{person}{Jure Leskovec}, {and} \bibinfo{person}{Stefanie Jegelka}.} \bibinfo{year}{2018}\natexlab{}.
\newblock \showarticletitle{How powerful are graph neural networks?}
\newblock \bibinfo{journal}{\emph{arXiv preprint arXiv:1810.00826}} (\bibinfo{year}{2018}).
\newblock


\bibitem[Yan et~al\mbox{.}(2020)]%
        {tinygnn}
\bibfield{author}{\bibinfo{person}{Bencheng Yan}, \bibinfo{person}{Chaokun Wang}, \bibinfo{person}{Gaoyang Guo}, {and} \bibinfo{person}{Yunkai Lou}.} \bibinfo{year}{2020}\natexlab{}.
\newblock \showarticletitle{Tinygnn: Learning efficient graph neural networks}. In \bibinfo{booktitle}{\emph{Proceedings of the 26th ACM SIGKDD International Conference on Knowledge Discovery \& Data Mining}}. \bibinfo{pages}{1848--1856}.
\newblock


\bibitem[Yang et~al\mbox{.}(2021)]%
        {cpf}
\bibfield{author}{\bibinfo{person}{Cheng Yang}, \bibinfo{person}{Jiawei Liu}, {and} \bibinfo{person}{Chuan Shi}.} \bibinfo{year}{2021}\natexlab{}.
\newblock \showarticletitle{Extract the Knowledge of Graph Neural Networks and Go Beyond It: An Effective Knowledge Distillation Framework}. In \bibinfo{booktitle}{\emph{Proceedings of the Web Conference 2021}} (Ljubljana, Slovenia) \emph{(\bibinfo{series}{WWW '21})}. \bibinfo{publisher}{Association for Computing Machinery}, \bibinfo{address}{New York, NY, USA}, \bibinfo{pages}{1227–1237}.
\newblock
\showISBNx{9781450383127}
\urldef\tempurl%
\url{https://doi.org/10.1145/3442381.3450068}
\showDOI{\tempurl}


\bibitem[Yang et~al\mbox{.}(2013)]%
        {ensemble1}
\bibfield{author}{\bibinfo{person}{Jing Yang}, \bibinfo{person}{Xiaoqin Zeng}, \bibinfo{person}{Shuiming Zhong}, {and} \bibinfo{person}{Shengli Wu}.} \bibinfo{year}{2013}\natexlab{}.
\newblock \showarticletitle{Effective neural network ensemble approach for improving generalization performance}.
\newblock \bibinfo{journal}{\emph{IEEE transactions on neural networks and learning systems}} \bibinfo{volume}{24}, \bibinfo{number}{6} (\bibinfo{year}{2013}), \bibinfo{pages}{878--887}.
\newblock


\bibitem[Yang et~al\mbox{.}(2022)]%
        {yang2022graph}
\bibfield{author}{\bibinfo{person}{Yaming Yang}, \bibinfo{person}{Ziyu Guan}, \bibinfo{person}{Wei Zhao}, \bibinfo{person}{Weigang Lu}, {and} \bibinfo{person}{Bo Zong}.} \bibinfo{year}{2022}\natexlab{}.
\newblock \showarticletitle{Graph substructure assembling network with soft sequence and context attention}.
\newblock \bibinfo{journal}{\emph{IEEE Transactions on Knowledge and Data Engineering}} \bibinfo{volume}{35}, \bibinfo{number}{5} (\bibinfo{year}{2022}), \bibinfo{pages}{4894--4907}.
\newblock


\bibitem[Yang et~al\mbox{.}(2020)]%
        {lsp}
\bibfield{author}{\bibinfo{person}{Yiding Yang}, \bibinfo{person}{Jiayan Qiu}, \bibinfo{person}{Mingli Song}, \bibinfo{person}{Dacheng Tao}, {and} \bibinfo{person}{Xinchao Wang}.} \bibinfo{year}{2020}\natexlab{}.
\newblock \showarticletitle{Distilling knowledge from graph convolutional networks}. In \bibinfo{booktitle}{\emph{Proceedings of the IEEE/CVF Conference on Computer Vision and Pattern Recognition}}. \bibinfo{pages}{7074--7083}.
\newblock


\bibitem[Zhang et~al\mbox{.}(2023)]%
        {zhang2023iterative}
\bibfield{author}{\bibinfo{person}{Hanlin Zhang}, \bibinfo{person}{Shuai Lin}, \bibinfo{person}{Weiyang Liu}, \bibinfo{person}{Pan Zhou}, \bibinfo{person}{Jian Tang}, \bibinfo{person}{Xiaodan Liang}, {and} \bibinfo{person}{Eric~P Xing}.} \bibinfo{year}{2023}\natexlab{}.
\newblock \showarticletitle{Iterative graph self-distillation}.
\newblock \bibinfo{journal}{\emph{IEEE Transactions on Knowledge and Data Engineering}} (\bibinfo{year}{2023}).
\newblock


\bibitem[Zhang et~al\mbox{.}(2019)]%
        {zhang2019your}
\bibfield{author}{\bibinfo{person}{Linfeng Zhang}, \bibinfo{person}{Jiebo Song}, \bibinfo{person}{Anni Gao}, \bibinfo{person}{Jingwei Chen}, \bibinfo{person}{Chenglong Bao}, {and} \bibinfo{person}{Kaisheng Ma}.} \bibinfo{year}{2019}\natexlab{}.
\newblock \showarticletitle{Be your own teacher: Improve the performance of convolutional neural networks via self distillation}. In \bibinfo{booktitle}{\emph{Proceedings of the IEEE/CVF International Conference on Computer Vision}}. \bibinfo{pages}{3713--3722}.
\newblock


\bibitem[Zhang et~al\mbox{.}(2021)]%
        {glnn}
\bibfield{author}{\bibinfo{person}{Shichang Zhang}, \bibinfo{person}{Yozen Liu}, \bibinfo{person}{Yizhou Sun}, {and} \bibinfo{person}{Neil Shah}.} \bibinfo{year}{2021}\natexlab{}.
\newblock \showarticletitle{Graph-less Neural Networks: Teaching Old MLPs New Tricks Via Distillation}. In \bibinfo{booktitle}{\emph{International Conference on Learning Representations}}.
\newblock


\bibitem[Zhang et~al\mbox{.}(2020)]%
        {rdd}
\bibfield{author}{\bibinfo{person}{Wentao Zhang}, \bibinfo{person}{Xupeng Miao}, \bibinfo{person}{Yingxia Shao}, \bibinfo{person}{Jiawei Jiang}, \bibinfo{person}{Lei Chen}, \bibinfo{person}{Olivier Ruas}, {and} \bibinfo{person}{Bin Cui}.} \bibinfo{year}{2020}\natexlab{}.
\newblock \showarticletitle{Reliable Data Distillation on Graph Convolutional Network}. In \bibinfo{booktitle}{\emph{Proceedings of the 2020 ACM SIGMOD International Conference on Management of Data}} (Portland, OR, USA) \emph{(\bibinfo{series}{SIGMOD '20})}. \bibinfo{publisher}{Association for Computing Machinery}, \bibinfo{address}{New York, NY, USA}, \bibinfo{pages}{1399–1414}.
\newblock
\showISBNx{9781450367356}
\urldef\tempurl%
\url{https://doi.org/10.1145/3318464.3389706}
\showDOI{\tempurl}


\bibitem[Zhu et~al\mbox{.}(2003)]%
        {zhu2003semi}
\bibfield{author}{\bibinfo{person}{Xiaojin Zhu}, \bibinfo{person}{Zoubin Ghahramani}, {and} \bibinfo{person}{John~D Lafferty}.} \bibinfo{year}{2003}\natexlab{}.
\newblock \showarticletitle{Semi-supervised learning using gaussian fields and harmonic functions}. In \bibinfo{booktitle}{\emph{Proceedings of the 20th International conference on Machine learning (ICML-03)}}. \bibinfo{pages}{912--919}.
\newblock


\end{thebibliography}

\appendix
\section{Implement Details}
\label{app:imp}
\para{Hyper-parameters.}
We set the max training epochs at $500$ for all the trails. The search space of the hyper-parameters is as follows:
\begin{itemize}
	\item Hidden Dimensionality $F = \{128, 256, 512, 1024, 2048\}$
	\item Number of Layer $L = \{2, 3\}$
	\item Ensemble Size $K = \{2, 3\}$
	\item Balance Parameter $\lambda, \lambda_{\mathrm{NA}} = \{0.1, 0.2, \cdots, 0.9\}$
	\item Divergence Sensitivity Parameter $\beta = \{0.5, 1, 2, 3, 4\}$ 
\end{itemize}
For masking rate $\rho$, we fix it at $0.1$ in the normal setting and set to the same value as the feature missing rate in the incomplete test data setting. 

\para{Hardware and Software.}
\adagmlp~is implemented based on the DGL library~\cite{dgl} and PyTorch 1.7.1 with Intel(R) Core(TM) i9-10980XE CPU @ 3.00GHz and 2 NVIDIA TITAN RTX GPUs.

\begin{table*}[!htbp]
\begin{center}
\caption{Performance comparison with G2G methods.}
\vspace{-0.5em}
\label{tab:g2g}
\resizebox{\textwidth}{!}{
\begin{tabular}{c|ccccccc|c}

\toprule
\textbf{Method} & \texttt{Cora} & \texttt{Citeseer} & \texttt{Pubmed} & \texttt{Photo} & \texttt{CS} & \texttt{Physics} & \texttt{ogbn-arxiv} & $\bar{\Delta}_{GCN}$ \\ \midrule
MLPs & $56.66_{\pm2.02}$ & $59.88_{\pm0.59}$ & $71.94_{\pm1.24}$ & $78.16_{\pm2.76}$ & $87.17_{\pm1.04}$ & $87.24_{\pm0.61}$ & $53.60_{\pm1.31}$ & -\\ 
GCN & $ 82.02_{\pm0.98} $ & $71.88_{\pm0.34}$ & $77.24_{\pm0.23}$ & $90.60_{\pm2.15}$ & $89.73_{\pm0.67}$ & $92.29_{\pm0.58}$ & $ 71.22_{\pm0.18} $ & - \\ \midrule

LSP & $82.70_{\pm0.43}$ & $72.68_{\pm0.62}$ & $80.86_{\pm0.50}$ & $91.74_{\pm1.42}$ & $92.56_{\pm0.45}$ & $92.85_{\pm0.46}$ & $71.57_{\pm0.25}$ & $\uparrow 1.73 \%$ \\
GNN-SD & $82.54_{\pm0.36}$ & $72.34_{\pm0.55}$ & $80.52_{\pm0.37}$ & $91.83_{\pm1.58}$ & $91.92_{\pm0.51}$ & $93.22_{\pm0.66}$ & $70.90_{\pm0.23}$ & $\uparrow 1.41 \%$\\
TinyGNN & $83.10_{\pm0.53}$ & $73.24_{\pm0.72}$ & $81.20_{\pm0.44}$ & $92.03_{\pm1.49}$ & $93.78_{\pm0.38}$ & $93.70_{\pm0.56}$ & $72.18_{\pm0.27}$ & $\uparrow 2.47 \%$ \\
RDD & $83.68_{\pm0.40}$ & $73.64_{\pm0.50}$ & $81.74_{\pm0.44}$ & $92.18_{\pm1.45}$ & $\textbf{94.20}_{\pm0.48}$ & $94.14_{\pm0.39}$ & $72.34_{\pm0.17}$ & $\uparrow 2.94 \%$ \\
FreeKD & $83.84_{\pm0.47}$ & $73.92_{\pm0.47}$ & $81.48_{\pm0.38}$ & $92.38_{\pm1.54}$ & $93.65_{\pm0.43}$ & $93.87_{\pm0.48}$ & $\textbf{72.50}_{\pm0.29}$ & $\uparrow 2.91 \%$ \\ 
AdaGMLP (ours)  & $ \textbf{84.26}_{\pm0.83} $ & $ \textbf{75.42}_{\pm0.39} $ & $ \textbf{81.88}_{\pm0.53} $ & $ \textbf{92.60}_{\pm0.37} $ & $ 93.79_{\pm0.33} $ & $\textbf{94.38}_{\pm0.27}$ & $71.45_{\pm0.10}$ & $\uparrow 3.28 \%$  \\
\bottomrule
\end{tabular}} \vspace{-1em}
\end{center}
\end{table*} 

\section{Algorithm of \adagmlp}
\label{app:algorithm}
The algorithm of our \adagmlp~is presented in Algorithm~\ref{alg:adagmlp}.

\begin{algorithm}[ht!]
\caption{\adagmlp~Algorithm (Transductive)}
\label{alg:adagmlp}
\begin{algorithmic}[1]
\STATE {\bfseries Input:} GNN teacher's output $\mathbf{Z}^{m}$, hyperparameters $\tau$, $\beta$, $\lambda$, and $\lambda_{\mathrm{NA}}$
\STATE Initialize Node weights $w_{i} = \frac{1}{N}$ for all $i \in \mathcal{V}$, combining weights $\alpha^{(k)} = 1$ for $k = 1, 2, \dots, K$
\FOR{$t = 1$ to $T$}  
\STATE \textcolor{gray}{// Student MLP training} 
    \FOR{$k = 1$ to $K$} 
    \STATE \textcolor{gray}{// Random Classification}
    	\STATE Sample labeled nodes to obtain a subset $\labeledNodeSet_{k}$
        \STATE Train MS$_{k}$ with $\labeledNodeSet_{k}$ via RC objective Eq.~(\ref{eq:adagmlp-ce})
    \STATE \textcolor{gray}{// Node Alignment}
        \STATE Randomly mask features of nodes in $\labeledNodeSet_{k}$ 
        \STATE Train MS$_{k}$ using nodes with completed and partially masked features via NA objective Eq.~(\ref{eq:adagmlp-na})
    \STATE \textcolor{gray}{// AdaBoost Knowledge Distillation}
        \STATE Train MS$_{k}$ with $\mathbf{Z}^{m}$ and $w^{(k)}$ via AdaKD objective Eq.~(\ref{eq:adagmlp-kl})
        \STATE Calculate the error $e^{(k)}$ via Eq.~(\ref{eq:err})
        \STATE Calculate the combining weight $\alpha^{(k)}$ via Eq.~(\ref{eq:alpha})
        \STATE Update node weights $w_{i}$ via Eq.~(\ref{eq:update-w})
    \ENDFOR
    \STATE Normalize node weights $w_{i} \leftarrow \frac{w_{i}}{\sum_{i=1}^{N} w_{i}}$
\ENDFOR
\STATE Obtain final prediction $p_{i}$ for node $i$ via Eq.~(\ref{eq:pred})
\end{algorithmic}
\end{algorithm}

\begin{table*}[!htbp]
\begin{center}
\caption{Performance comparison with Ensemble methods.}
\vspace{-0.5em}
\label{tab:ense}
\resizebox{\textwidth}{!}{
\begin{tabular}{c|ccc|ccc|c}

\toprule
\textbf{Ensemble Method} & \multicolumn{3}{c|}{\texttt{Cora}} &  \multicolumn{3}{c|}{\texttt{Citeseer}} & $\bar{\Delta}_{GLNN}$ \\ 
\midrule

\multicolumn{8}{c}{\textbf{\texttt{Transductive Setting}}} \\

\midrule

None (GLNN) & \multicolumn{3}{c|}{$82.08_{\pm1.14}$} & \multicolumn{3}{c|}{$73.46_{\pm0.47}$} & - \\
Average & \multicolumn{3}{c|}{$81.62_{\pm0.97}$} & \multicolumn{3}{c|}{$72.12_{\pm0.40}$} & $\downarrow 1.19\%$ \\
Vote & \multicolumn{3}{c|}{$82.04_{\pm1.17}$} & \multicolumn{3}{c|}{$72.44_{\pm0.37}$} & $\downarrow 0.71\%$ \\
Bagging & \multicolumn{3}{c|}{$82.68_{\pm0.92}$} & \multicolumn{3}{c|}{$73.06_{\pm0.52}$} & $\uparrow 0.09\%$ \\
\adagmlp & \multicolumn{3}{c|}{$\textbf{84.26}_{\pm0.82}$} & \multicolumn{3}{c|}{$\textbf{75.42}_{\pm0.39}$} & $\uparrow \textbf{2.66}\% $ \\

\midrule

\multicolumn{8}{c}{\textbf{\texttt{Insufficient Training Data Setting}}} \\

\midrule
Label Rate & 1\% & 2\% & 3\% & 1\% & 2\% & 3\% & ~ \\
\midrule

None (GLNN) & $63.24_{\pm5.94}$ & $72.48_{\pm4.68}$ & $74.31_{\pm4.46}$ & $62.74_{\pm4.38}$ & $65.54_{\pm6.39}$ & $69.78_{\pm3.71}$ & - \\
Average & $62.36_{\pm5.97}$ & $72.68_{\pm2.72}$ & $75.26_{\pm1.64}$ & $57.16_{\pm6.92}$ & $64.12_{\pm2.49}$ & $67.28_{\pm1.08}$ & $\downarrow 2.14\%$ \\ 
Vote & $61.40_{\pm5.53}$ & $73.90_{\pm1.99}$ & $76.94_{\pm2.48}$ & $56.36_{\pm7.34}$ & $63.74_{\pm2.48}$ & $67.20_{\pm0.75}$ & $\downarrow 2.07\%$ \\
Bagging & $61.98_{\pm5.72}$ & $74.40_{\pm1.73}$ & $77.28_{\pm2.51}$ & $56.88_{\pm6.70}$ & $64.06_{\pm2.28}$ & $68.03_{\pm0.94}$ & $\downarrow 1.31\%$ \\
\adagmlp & $\textbf{71.20}_{\pm4.22}$ & $\textbf{77.92}_{\pm1.22}$ & $\textbf{80.38}_{\pm1.05}$ & $\textbf{63.84}_{\pm2.13}$ & $\textbf{66.46}_{\pm2.61}$ & $\textbf{69.96}_{\pm1.91}$ & $\uparrow \textbf{5.58}\%$ \\

\midrule

\multicolumn{8}{c}{\textbf{\texttt{Incomplete Test Data Setting}}} \\

\midrule
Feature Missing Rate & 10\% & 30\% & 50\% & 10\% & 30\% & 50\% & ~ \\
\midrule

None (GLNN) & $68.84_{\pm4.71}$ & $65.70_{\pm3.26}$ & $59.34_{\pm4.07}$ & $60.06_{\pm4.04}$ & $59.12_{\pm4.13}$ & $57.02_{\pm4.10}$ & - \\
Average & $71.90_{\pm0.82}$ & $68.00_{\pm0.56}$ & $61.94_{\pm0.82}$ & $61.34_{\pm0.41}$ & $60.30_{\pm0.90}$ & $58.10_{\pm0.61}$ & $\uparrow 3.05\%$ \\ 
Vote & $71.82_{\pm0.58}$ & $67.98_{\pm0.43}$ & $61.92_{\pm0.92}$ & $61.02_{\pm0.21}$ & $59.86_{\pm0.56}$ & $57.84_{\pm0.54}$ & $\uparrow 2.73\%$ \\
Bagging & $71.94_{\pm1.10}$ & $68.00_{\pm0.58}$ & $62.03_{\pm1.14}$ & $61.28_{\pm0.09}$ & $59.99_{\pm0.47}$ & $57.98_{\pm0.57}$ & $\uparrow 2.95\%$ \\
\adagmlp & $\textbf{82.42}_{\pm0.72}$ & $\textbf{80.46}_{\pm1.32}$ & $\textbf{78.54}_{\pm1.38}$ & $\textbf{73.14}_{\pm0.57}$ & $\textbf{71.16}_{\pm1.73}$ & $\textbf{64.36}_{\pm1.45}$ & $\uparrow \textbf{21.59}\%$ \\

\bottomrule
\end{tabular}} \vspace{-1em}
\end{center}
\end{table*} 

%
%

\begin{table}[!htbp]
\begin{center}
\caption{Datasets Statics.}
\vspace{0.5em}
\label{tab:data_stat}
\resizebox{\columnwidth}{!}{
\begin{tabular}{lccccc}
\toprule
\textbf{Dataset} & \textbf{\# Nodes} & \textbf{\# Edges} & \textbf{\# Features} & \textbf{\# Classes} & \textbf{Label Rate} \\ \midrule
\texttt{Cora} & 2,708 & 5,278 & 1,433 & 7 & 5.2\% \\
\texttt{Citeseer} & 3,327 & 4,614 & 3,703 & 6 & 3.6\% \\
\texttt{Pubmed} & 19,717 & 44,324 & 500 & 3 & 0.3\% \\
\texttt{Photo} & 7,650 & 119,081 & 745 & 8 & 2.1\% \\
\texttt{CS} & 18,333 & 81,894 & 6,805 & 15 & 1.6\% \\
\texttt{Physics} & 34,493 & 247,962 & 8,415 & 5 & 0.3\% \\
\texttt{ogbn-arxiv} & 169,343 & 1,166,243 & 128 & 40 & 53.7\% \\ \bottomrule
\end{tabular}} 
\end{center}
\end{table}

\section{Dataset Statics}
\label{app:data_stat}
Table~\ref{tab:data_stat} presents a summary of the statistical characteristics of these datasets. Data splitting strategies differ depending on the dataset's scale:
\begin{itemize}
	\item For the three small-scale datasets, \texttt{Cora}, \texttt{Citeseer}, and \texttt{Pubmed}, we adopt the data splitting strategy outlined in \cite{gcn}.
	\item For \texttt{Coauthor-CS}, \texttt{Coauthor-Physics}, and \texttt{Amazon-Photo}, we follow the procedures from \cite{glnn,cpf} to perform random data splits into training, validation, and test sets. 
	\item For the large-scale dataset, \texttt{ogbn-arxiv}, we strictly follow the  publicly available data splits in~\cite{ogb}.
\end{itemize}

\section{Performance Comparison with G2G Methods}
\label{app:comp-g2g}

We compare our \adagmlp~with SOTA GNN-to-GNN (G2G) methods, i.e., CPF~\cite{cpf}, RDD~\cite{rdd}, TinyGNN~\cite{tinygnn}, GNN-SD~\cite{gnn-sd}, and LSP~\cite{lsp}, in Table~\ref{tab:g2g}. All the methods use GCN as the teacher model. We reuse the results of G2G methods from~\cite{krd}. $\bar{\Delta}_{GCN}$ is the average improvements across all the datasets over GCN.

We can observe that \adagmlp~consistently shows better performance across the majority of the datasets. The improvements in accuracy are most notable in the \texttt{Cora}, \texttt{Citeseer}, and \texttt{Pubmed} datasets. On \texttt{Coauthor-CS} and \texttt{ogbn-arxiv}, \adagmlp~still demonstrates competitive performance, although not the top performer. It maintains robust results but with a slightly lower margin compared to the top performer (RDD and FreeKD).

\adagmlp~not only enhances efficiency but also maintains competitive accuracy. It achieves higher accuracy than G2G methods across multiple datasets. Unlike G2G methods, which require message propagation during inference, \adagmlp~operates without this need. This efficiency is crucial in real-world applications, especially in scenarios with latency constraints and resource limitations, making \adagmlp~an optimal choice for such settings. This balance between efficiency and accuracy is a significant advantage for practical applications where both factors are essential.

\section{Performance Comparison with Ensemble Methods}
\label{app:comp-ense}
In this section, we compare our \adagmlp against some two well-known ensemble strategies. i.e., Vote, Bagging~\cite{ensemble}, and a simple average ensemble strategy that uses the average predictions from each MLP student. All the strategies use the same configuration. We conduct experiments under three different settings, including the transductive setting, insufficient training data setting, and incomplete test data setting on Cora and Citeseer. The results are provided in Table~\ref{tab:ense}

In the transductive setting, our AdaBoost strategies achieve the highest accuracy on both the Cora and Citeseer datasets. Other strategies, such as average, vote, and bagging, perform relatively close to the baseline GLNN method but fall short of surpassing AdaBoost. This is attributed to AdaBoost's adaptive weighting of each student and emphasis on unaligned knowledge points, allowing it to focus on the difficult-to-extract knowledge and improve overall predictive performance.

In the insufficient training data setting, we can see that simple ensemble strategies can also achieve better performance compared to GLNN in some cases. However, there is still a performance gap between them and AdaBoost. It indicates that the AdaBoost's ability to adaptively weigh weak learners is particularly effective in tackling the challenges posed by limited labeled data.

In the incomplete test data setting, we can observe that simple ensemble strategies can also bring performance improvement when test data is corrupted. It demonstrates that combining multiple MLP students is a promising and simple way to mitigate the incomplete test data issue.

Overall, the results show that AdaBoost outperforms other ensemble strategies in various settings. Its adaptability, emphasis on challenging knowledge points, and weighting mechanism contribute to its superior performance. Additionally, the experiments highlight the potential benefits of ensemble methods for improving performance of G2M.

\end{document}